\documentclass[journal]{IEEEtran}
\usepackage{graphicx}
\pdfoutput=1
\usepackage[linesnumbered,ruled,lined]{algorithm2e}
\usepackage{amssymb}
\usepackage{amsmath}
\usepackage[left,pagewise]{lineno}
\usepackage{colortbl,booktabs}
\usepackage{threeparttable}
\usepackage{setspace}
\usepackage{bm}
\usepackage{url}
\usepackage{subfigure}

\usepackage{xcolor}
\usepackage{multirow}

\def\BibTeX{{\rm B\kern-.05em{\sc i\kern-.025em b}\kern-.08em
    T\kern-.1667em\lower.7ex\hbox{E}\kern-.125emX}}
\usepackage{balance}

\begin{document}

\title{Beyond Sharing Weights in Decoupling Feature Learning Network for UAV RGB-Infrared Vehicle Re-Identification}
% \title{UAV RGB-Infrared Cross-Modality Vehicle Re-Identification}

\author{\IEEEauthorblockN{Xingyue~Liu, Jiahao~Qi, Chen~Chen, Kangcheng~Bin and Ping~Zhong~\IEEEmembership{Senior Member,~IEEE}}
\thanks{This work was supported in part by Natural Science Foundation of China under Grant 61971428.\emph{(Corresponding author: Ping Zhong)}}
\thanks{Xingyue Liu, Jiahao Qi,  Chen Chen, Kangcheng~Bin and Ping Zhong are with the National Key Laboratory of Automatic Target Recognition, National University of Defense Technology, Changsha 410073, China (e-mail: liuxingyue18@nudt.edu.cn, qijiahao1996@nudt.edu.cn, chenchen21c@nudt.edu.cn, binkc21@nudt.edu.cn, zhongping@nudt.edu.cn).}
}

\markboth{Submitted to IEEE Transactions on Multimedia}%
{Liu \MakeLowercase{\textit{et al.}}: orientation-Invariant Feature Learning for UAV RGB-Infrared Cross-Modality Vehicle Re-Identification}

\maketitle

\begin{abstract}
% Vehicle re-identification (Re-ID) with unmanned aerial vehicles (UAV) is gaining more attention owing to its wide applications in video surveillance and public security.
% While, most works is based on RGB Re-ID data without considering the limitation of poor illumination.
% As infrared imaging is famous for its robustness to illumination which could offer complementary information.
% Hence, cross-modality UAV vehicle Re-ID with RGB and IR images could further promotes the applications of UAV vehicle Re-ID, which makes it possible to perform full-time target searching.
Owing to the capacity of performing full-time target search, cross-modality vehicle re-identification based on unmanned aerial vehicle (UAV) is gaining more attention in both video surveillance and public security. 
However, this promising and innovative research has not been studied sufficiently due to the data inadequacy issue.
Meanwhile, the cross-modality discrepancy and orientation discrepancy challenges further aggravate the difficulty of this task. 
To this end, we pioneer a cross-modality vehicle Re-ID benchmark named UAV Cross-Modality Vehicle Re-ID (UCM-VeID), containing 753 identities with \textbf{16015} RGB and \textbf{13913} infrared images.
% To make full use of the proposed dataset, we further annotate the color attributes of vehicles.
Moreover, to meet cross-modality discrepancy and orientation discrepancy challenges, we present a hybrid weights decoupling network (HWDNet) to learn the shared discriminative orientation-invariant features. 
For the first challenge, we proposed a hybrid weights siamese network with a well-designed weight restrainer and its corresponding objective function to learn both modality-specific and modality shared information.
In terms of the second challenge, three effective decoupling structures with two pretext tasks are investigated to flexibly conduct OIFS task.
% In addition, a novel loss function is presented for cross-modality orientation invariance learning.
Comprehensive experiments are carried out to validate the effectiveness of the proposed method.
The dataset and codes will be released at \url{https://github.com/moonstarL/UAV-CM-VeID}.
\end{abstract}


\section{Introduction} \label{sec:1}
\IEEEPARstart{U}{nmanned} aerial vehicles (UAV) vehicle Re-ID aims to match the given vehicles in gallery dataset collected by different UAV platforms with non-overlapping views.
Owing to the demand of video surveillance and social security \cite{9162561, DBLP, 10209282,9019850}, UAV vehicle Re-ID has been drawing growing attention in both industry and academia \cite{2016Matthias}.
Recent efforts in the UAV vehicle Re-ID have been devoted to single modality (RGB), which demonstrated limited performances under poor illumination condition.
More specifically, little credible appearance of targets can be captured by RGB in dark or over-exposure environments.
In this case, infrared (IR) camera shows great superiority to tackle the above issue, since it can describe the credible appearance of targets with the help of thermal radiation, instead of the external lighting.
Then, full-time UAV Re-ID can be realized by taking full advantage of detailed information (rich texture and color information) of RGB cameras and night vision capability of infrared cameras.
Given an UAV RGB (IR) vehicle image, the task, named \emph{cross-modality vehicle Re-ID}, purposes to find the corresponding IR (RGB) vehicle image in the gallery dataset.
% For convenience, this task is abbreviated as \textbf{cross-modality vehicle Re-ID} in the rest of this paper. 
\begin{figure}[!t]                  
  \centering                    
  \includegraphics[width=\linewidth]{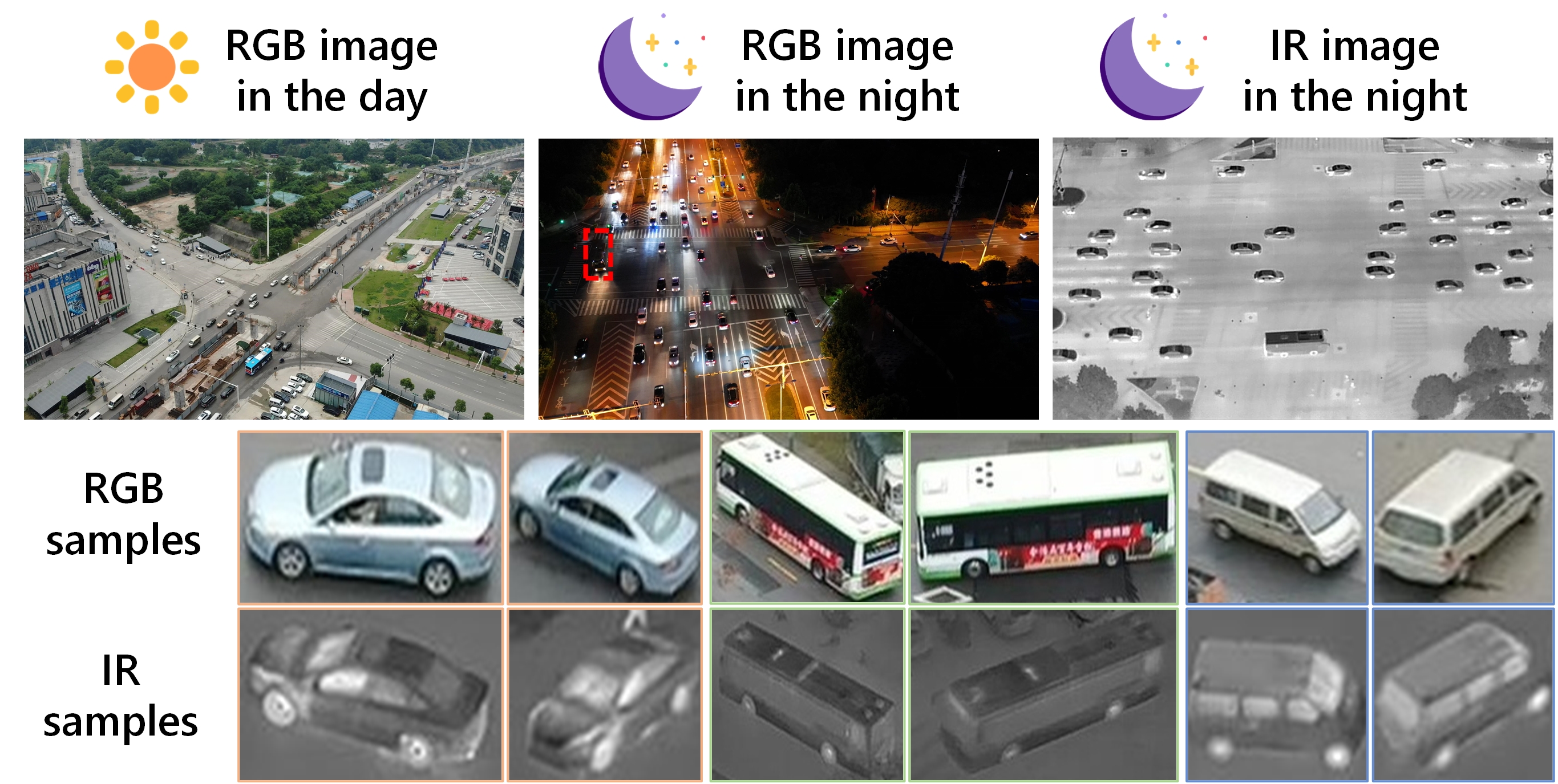}                  
  \caption{Examples of RGB samples and IR samples captured in the day and night. The pictures in the first row show the data acquisition scene. The second row and the third row are RGB samples and IR, respectively. Samples with same color in the last two rows represent the same vehicle.}                  
  \label{fig:C1}    
\end{figure}

To the best of our knowledge, UAV cross-modality vehicle Re-ID still remains an under-explored task, and the existing datasets are inadequate to support this task research.
Precisely, the public UAV vehicle Re-ID datasets, such as VRAI \cite{VRAI}, UAV-VeID \cite{UAV-VeID}, VeRi-UAV \cite{VeRi-UAV} and VRU \cite{VRU}, only collect the RGB modal data while lacking of the corresponding IR modal data.
At the same time, the multi-spectral vehicle Re-ID datasets (RGBN300, RGBNT100 \cite{RGBN300} and MSVR310 \cite{MSVR310}) are limited by their fixed data collection views and complicated data preprocessing process.
Therefore, a comprehensive and public dataset is imperative in the cross-modality vehicle Re-ID community.

To satisfy this urgent demand of research data, a diversified and longitudinal dataset, named UAV cross-modality vehicle Re-ID (UCM-VeID), is proposed in this paper. 
UCM-VeID contains 29939 vehicles of 753 identities, including 16015 RGB samples and 13913 IR samples, as displayed in Fig. \ref{fig:C1}.
Meanwhile, the targets in UCM-VeID possess the multiple-view and multiple-scale characteristics, which further adds to the difficulty of cross-modality vehicle Re-ID task.
To further explore the dataset, 8 orientation classes, 9 vehicle types and 12 vehicle colors are annotated to make full use of the appearance information.
Moreover, as far as we known, UCM-VeID refers to the first open-source UAV cross-modality benchmark in the vehicle Re-ID community. 
Last but not the least, UCM-VeID also has a good potential to benefit other researches that face the similar challenges with heterogeneous modalities, changeable views, variable scales.
% RGB and IR images are heterogeneous modality data with different data channels, due to the distinct image forming principles and wavelength coverage between RGB and IR cameras.

\begin{figure}[!t]
  \centering
  \subfigure[]{\label{fig:24_a}\includegraphics[width=0.88\linewidth]{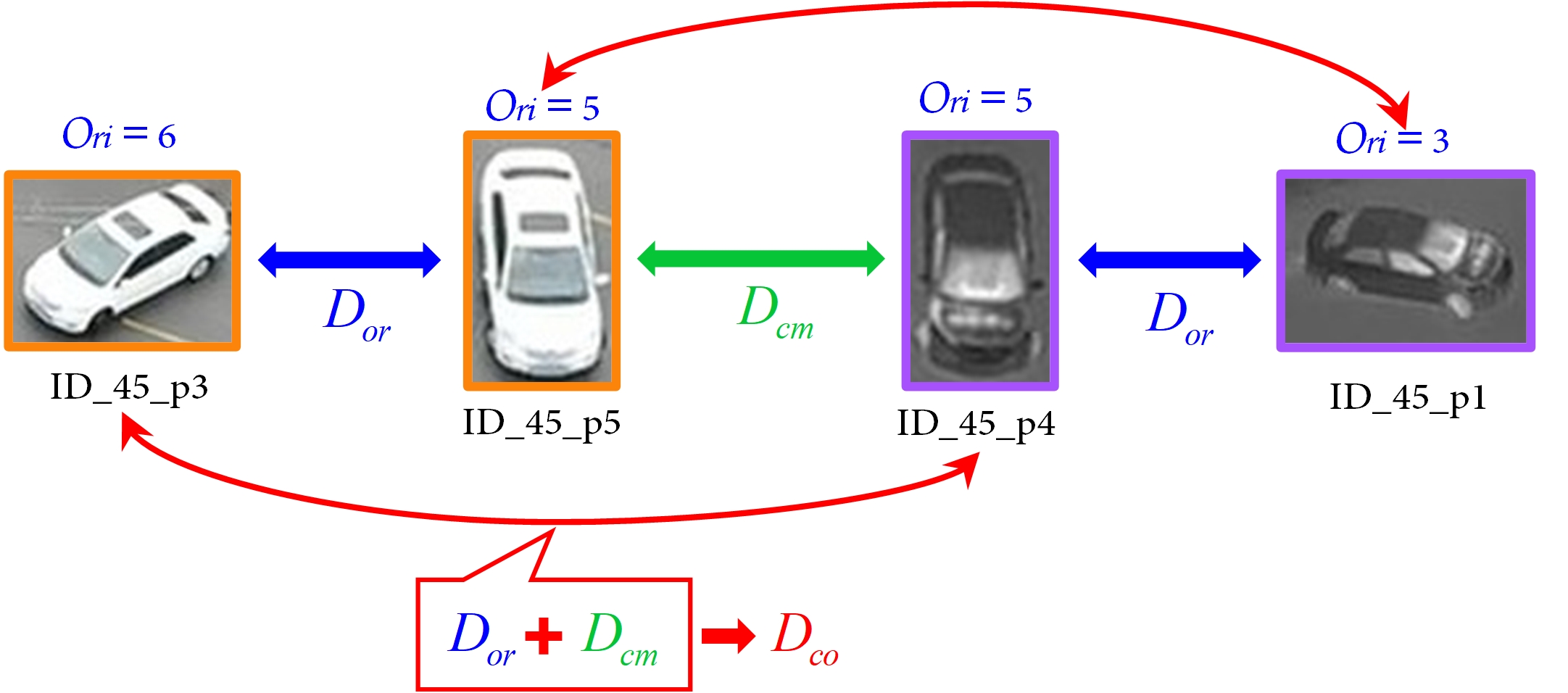}}
  \subfigure[]{\label{fig:24_b}\includegraphics[width=0.88\linewidth]{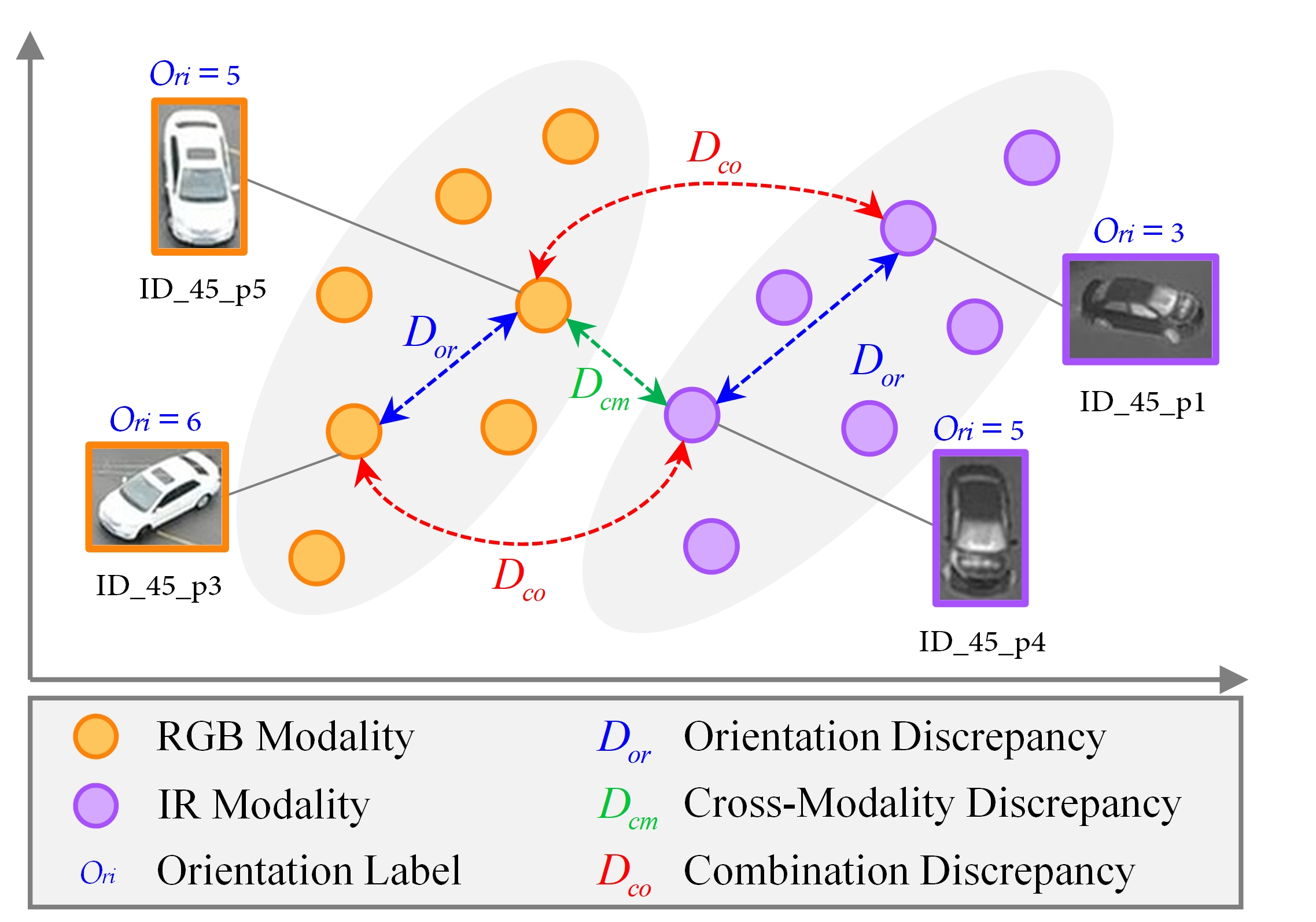}}
  %\vspace{-2ex}
  \caption{Illustration of the key challenges in the UAV cross-modality vehicle Re-ID task. (a) The illustration of challenge from appearance aspect, (b) The illustration of challenge from representation aspect. This diagram shows the different modal samples of the same ID in visual and embedding space. The orientation discrepancy suffers from different views with UAV platform and the cross-modality discrepancy are due to different modalities cameras. Combination discrepancy contains both orientation discrepancy and cross-modality discrepancy.}
  \label{fig:C2}
\end{figure}

Apart from the problem of insufficient data, there still remain some technical challenges in cross-modality vehicle Re-ID task. 
The representative one is the image misalignment between different samples with same ID, as illustrated in Fig. \ref{fig:C2}.
To be specific, the image misalignment is mainly arose by the \textbf{cross-modality discrepancy} and \textbf{orientation discrepancy}.
Firstly, cross-modality discrepancy mainly exists in inter-modality samples, as shown by ID\_45\_p4 and ID\_45\_p5 in Fig. \ref{fig:C2}, which is derived from the differences of imaging principle and imaging wavelength between RGB and IR sensors \cite{AncongWu2017, LiuTZ21}.
This discrepancy makes it difficult to extract the unified latent representation for the inter-modality samples.
As for the orientation discrepancy, it describes the appearance difference of intra-modality samples in the diverse views of UAV platform, as depicted by ID\_45\_p3 and ID\_45\_p5 in Fig. \ref{fig:C2}. 
It may mislead a deep learning model to take orientation-relevant features as discriminative cues for Re-ID task.
Even worse, the interaction between cross-modality discrepancy and orientation discrepancy, named combination discrepancy, would further exacerbate the image misalignment challenge, as presented with ID\_45\_p3 and ID\_45\_p4 in Fig. \ref{fig:C2}.

For the sake of addressing the aforementioned challenges, a large number of works have been reported in the past decades.
To eliminate the cross-modality discrepancy, the mainstream works employed a two-stream network, which embeds RGB and IR inputs into a shared latent space by sharing network parameters to extract modality-shared features \cite{IJCVAncongWu, 2019YiHao, WuD0LWHZJ21, ZhangLX21, YeSS21, WangZSZWYL21, HuangWXZZZ22}.
% While imposing sharing network parameters is deficient in learning discriminant features due to the inadequate use of the modality-specific information.
Besides, to take both modality-specific and modality-shared information into consideration, most methods \cite{YeWLY18, HaoWLG19, Bi-Directional2020, YeCSS22, 2020Ensemble, LiuXJ23, 2023Yukang,HomogeneousYe} manually designed various two-stream network structures to keep the shallow layer parameters distinct and the deep layer parameters shared, where deep layers and shallow layers are set up differently.
% Furthermore, Liu \emph{et al.} \cite{LiuTZ21} and Fu \emph{et al.} \cite{FuH0S0H21} explored the best way to balance shared deep layer parameters with distinct shallow layer parameters, obtaining the optimal network parameters sharing scheme with resnet50.
% The modality-specific features can be learned by distinct shallow layers still have large cross-modality discrepancy, which is not conducive to the subsequent learning process of deep shared networks. 
% As a result, shared features which would have been learned in shallow layers might be lost.
The purpose of the above designed networks is to extract both discriminative and shared features.
Distinct shallow layers can extract modality-specific information, such as color and intensity information, which enhance the discrimination of the representation.
However, it may lose some of the low-level shared semantic at the same time, which is not conducive to the subsequent learning process of deep shared networks.
% Furthermore, Liu \emph{et al.} \cite{} obtained the optimal network parameter sharing scheme with resnet50 as the backbone through massive experiments.
% Fu \emph{et al.} \cite{} observed that sharing the BN layer is the key to improving the performance of cross-modality Re-ID and based on this result, proposed Cross-Modality Neural Architecture Search method, to automatically search the optimal separation scheme for BN layers.

In terms of orientation discrepancy, many existing works focus on learning orientation-invariant features, which are robust to diversity of target orientations.
These researches can be roughly divided into three categories. 
In the first category, the methods \cite{SunNXY20,YangCZSLM22,LiLZHZ23,10105456,9248611} tried to extract some local detail feature maps as the reference information to specify different orientations. 
% He \emph{et al.} 
The methods \cite{VeRi-UAV,ChenLWC20,2018Zhouyi,MengLLLYZGWH20,LiuLZY020} in the second category used extra semantic information (such as key-points, mask) as the supervised signal to assist the deep learning models to acquire orientation-invariant features. 
When it comes to the methods in the third category \cite{0001LLPG21,LinLYPX19}, they employ some unsupervised strategies to establish the relationship between the representation of different orientations, which makes the model adapt the variation of orientations. 
Compared with the above delicate methods, a straightforward and effective methodology named feature decoupling is widely used to obtain the orientation-invariant features. 
Bai \emph{et al.} \cite{BaiLLWD22} employed this methodology to design a specific disentangled feature learning network for vehicle Re-ID. 
% Following this work, in this paper, we also adopt the feature decoupling conception to address the orientation discrepancy
% Meanwhile, compared with fixed surveillance cameras, UAV vehicle Re-ID is more challenge because of changeable and special view-points, such as bird-eye view.
% While gathering more comprehensive target information, UAV cameras also records some task-agnostic information, such as target orientation.
% Samples of same vehicle show distinct orientations owing to various camera views, as illustrated in Fig. \ref{fig:C1}.
% The variable orientations leads to a large intra-class differences.
% What's worse, cross-modal data further exacerbates the challenges posed by variable orientations, due to the difference between RGB and IR data characteristics.
% While most existing methods focus on learning discriminant features without taking orientation into account, resulting that models mistakenly use orientation-relevant features as discriminants for Re-ID task.
% As confirmed in Fig. \ref{fig:C2}, the gallery samples that have the identical or symmetrical orientation with the given query sample are selected first, regardless of whether they are correct or not.

There is no doubt that significant effort have been made by the existing methods to tackle the cross-modality and orientation discrepancies. 
However, there exists no research to address the problem combination problem of these two discrepancies in an end-to-end fashion.
Unfortunately, as stated beforehand, this combination would further aggravate the difficulty of cross-modality vehicle Re-ID task. 
To this end, we propose a hybrid weights decoupling network (HWDNet) to tackle the combination of cross-modality discrepancy and orientation discrepancy issues.
First of all, we introduce a hybrid weights siamese network based on a two-stream network with a well-designed weight restrainer.
The intuition behind the weight restrainer is straightforward: there is a relation between RGB and IR images when they describe the same target, even if they are heterogeneous modality data.
That is, the basic semantic information of RGB and IR images for the same target are different but related.
Correspondingly, the relationship of two-stream network weights should also be related, instead of specific or shared. 
% For the traditional where the shallow layers learn modality-specific representations and the deep layers extract modality-invariant features.
For this purpose, a weight restrainer is designed to employ a linear weight loss used for constraining the weights of shallow layers.
In addition, we investigate an adaptive scheme to determine which layers should be related and which layers should be shared. 
\par
Secondly, the orientation discrepancy problem is solved with two operations: orientation-invariant feature separation (OIFS) and orientation-invariant feature refining (OIFR). 
For OIFS, we investigate the principle of decoupling strategy and three simple but efficient decoupling structure are devised to spilt the original feature representation into the orientation-invariant part and orientation-relevant part. 
In order to achieve a better decoupling result, the supervised pretext task with orientation labels is designed for restricting the orientation-relevant part on the basis of orientation classification. 
As for the OIFR, we introduce a statistical centroid guided pretext task in a self-supervised manner to further alleviate the effect of vehicle orientation for Re-ID task. 
Considering the cross-modality discrepancy existed, the statistical centroid is derived by simultaneously taking multi-modal feature representations into account.
% \par

The main contributions of this work can be briefly summarized as below.

(1) For the first time, we contribute a benchmark dataset named UCM-VeID to support the research of cross-modality vehicle Re-ID and propose HWDNet for UAV cross-modality vehicle Re-ID task.
% Compared with existing vehicle Re-ID datasets, UCM-VeID contains RGB and IR two different modal images without manual alignment, facing a more challenging but potential vehicle Re-ID task.

(2) Hybrid weights siamese network devotes to decrease cross-modality discrepancy by introducing a new structure named weight restrainer with a weight linear loss function between unshared shallow layers.

(3) To eliminate the orientation discrepancy, we figure out the most appropriate decoupling strategy and design two pretext tasks named orientation classification task and statistical centroid guided pretext task.
% Numerous experiments confirm that the method has promising performance on UCM-VeID and existing vehicle Re-ID datasets.

\section{Related Works}
In this section, we briefly review the related work with respect to vehicle Re-ID, cross-modality person Re-ID and orientation-invariant feature learning.
\par
\subsection{Vehicle Re-ID Datasets}
In recent years, vehicle Re-ID researches have been well developed along with the strong support of vehicle Re-ID datasets.
% The existing vehicle Re-ID datasets could be divided into three categories: based on surveillance cameras, based on UAV cameras and based on multi-spectral.
% \emph{1) based on surveillance cameras:}
Benefiting from the dense urban surveillance cameras, more and more high-quality vehicle Re-ID datasets have emerged.
2016, Liu \emph{et al.} \cite{2016Liu} proposed the first vehicle Re-ID dataset, named VehicleID, which is one of the main standard dataset for vehicle Re-ID community. 
The other one is VeRI-776 which is built up by Liu \emph{et al.} \cite{LiuLMM16} in the same year.
Subsequently, datasets such as VD1 and VD2 \cite{YanTWZH17}, VERI-Wild \cite{LouB0WD19} and Vehicle-1M \cite{GuoZLWL18} were proposed to further optimize and expand the vehicle Re-ID dataset in terms of view-points, camera number, time span, background complexity, precise model type and detail attribute annotations, respectively.
% The above mentioned datasets are based on fixed surveillance cameras and are also only applicable to urban surveillance scenarios.
Meanwhile, with the development of UAV, some vehicle Re-ID datasets based on UAV cameras are proposed, such as VRAI \cite{VRAI}, UAV-VeID \cite{UAV-VeID}, VRU \cite{VRU} and VeRi-UAV \cite{VeRi-UAV}.
Comparing with the fixed surveillance cameras, UAV-based cameras show the superiority of better mobility, flexibility and convenience.
As a result, UAV-based datasets have more complete target characteristics in terms of richer and special viewpoints, variable scales, and complex interference, enabling more effective and proactive vehicle Re-ID.
While, the above mentioned datasets are only based on RGB cameras, which has too poor performance at night to achieve full-time vehicle Re-ID.

To address the above issue, Li \emph{et al.} \cite{RGBN300} first proposed multi-spectral vehicle Re-ID problem with a baseline and released two benchmark datasets named RGBN300 and RGBNT100.
On this basis, Zheng \emph{et al.} \cite{MSVR310} presented a cross-directional consistency network and provided a RGB-NIR-TIR multi-spectral vehicle Re-ID benchmark MSVR310.
However, cross-modality vehicle Re-ID is different from multi-spectral vehicle Re-ID. 
The view angle of multi-spectral vehicle Re-ID data collection is fixed and approximates at eye level, resulting in a information focus on vehicle body side.
While UAV is more active and flexible which can overlook the targets. 
As a consequence, more effective targets information can be accessed quickly and efficiently. 
% Secondly, from Re-ID model perspective, multi-spectral vehicle Re-ID is designed to exploit the complementary characteristics between manually aligned images of different modalities which is time consuming and impractical to scale.
% Whereas, cross-modality vehicle Re-ID pays more attention on shared discriminant features of misaligned multi-modal images. 
% Thus, it is eliminates the need for cumbersome data processing so as to implement and promote.
% Hence cross-modality vehicle Re-ID is valuable to be explored.

\subsection{RGB-IR Cross-Modality Re-ID}
RGB-IR cross-modality person Re-ID task completes the matching of the same pedestrian images taken by different modality cameras.
Cross-modality shift is the main problem for RGB-IR cross-modality person Re-ID task.
To deal with the above mentioned problem, a approach is to generate missing modality-specific information according to the existing ones \cite{Hi-CMD, ThermalGAN, SFANet, 2020Xian,Syncretic-Modality}, aiming at reducing cross-modality discrepancy by aligning different modality data.
As a representative, Wang \emph{et al.} \cite{WangZYCCLH20} devised to generate cross-modality paired-images from existing RGB and IR images for modality alignment and Ye \emph{et al.} \cite{YeR0S21} utilized RGB images to generate an auxiliary modality to decrease the intra-class distance.
Although cross-modality Re-ID methods based on generation are effective, the training process is complex, tends to introduce noise and is difficult to converge.
% RGB-Infrared Person Re-Identification via Homogeneous Augmented Tri-Modal Learning
% RGB-Infrared Cross-Modality Person Re-Identification via Joint Pixel and Feature Alignment
% Cross-Modality Paired-Images Generation for RGB-Infrared Person Re-Identification
\par
The other approach which has a brief network structure and excludes generation process, devotes to learning a shared latent space to eliminate modality shift.
These methods are based on representation method and metric learning method, by designing feature extraction modal and metric loss function to direct shared discriminant feature learning from RGB and IR modalities.
According to the different model framework, there are two kinds of typical methods.
The first one called one-stream network \cite{IJCVAncongWu, 2019YiHao, WuD0LWHZJ21, ZhangLX21, YeSS21, WangZSZWYL21, HuangWXZZZ22,10179157}, shares the whole network parameters between RGB and IR modalities, which sacrifices discriminative feature learning for the sake of extracting modality invariant features.
Different from one-stream network, two-stream network uses unshared shallow layers to learn modality-specific features and shares deep layers for embedding single modality features into shared feature latent space\cite{YeWLY18, HaoWLG19, Bi-Directional2020, YeCSS22, 2020Ensemble, LiuXJ23, 2023Yukang}.
However, there is still a large cross-modality discrepancy in the pair of features learned in the unshared shallow layer.
More importantly, shared features are lost that may have been learned in the shallow layers.
This motivates us to design a new method which utilizes both modality-specific information and modality-shared information of shallow layers to enhance discriminability of features.
More concretely, we proposed a new structure called weight restrainer, where the unshared layers of two-stream network possess a linear relationship.
In addition, a weight linear loss function is introduced to direct the linear relationship of unshared parameters, for keeping modality-specific information while mining modality shared characteristics. 

\subsection{Orientation-Invariant Feature Embedding}
% Due to different view-points of cameras, one target contains various orientation samples in vehicle Re-ID datasets.
% Orientation variance leads to dramatically different appearances of the same vehicle, which make the vehicle Re-ID a challenge task.
% Therefore, it necessary to extract orientation-invariant feature, which is robust to the diversity of vehicle orientations.
For vehicle Re-ID task, orientation variance caused by different view-points, leads to dramatically different appearances of the same vehicle, which make the vehicle Re-ID a challenge task.
Several researches have been carried out to tackle variation view-points issue to learn orientation-invariant features.
Wang \emph{et al.} \cite{WangTLYYSYWLW17} and Khorramshahi \emph{et al.} \cite{Khorramshahi0PR19} utilize 20 key points annotations to generate view-related attention maps for acquiring orientation-based localized discriminative features.
Similar to the function of key points, pose mask \cite{MengLLLYZGWH20} and view segmentation \cite{LiuLZY020} are used to obtain latent orientation-relevant information.
However, the network suffers from semantic information loss during learning process guided by these semantic attribute labels, which are insufficient to cover all discriminant features.
And these semantic attributes labels (key points, pose mask and view segmentation) are annotated manually, which is time consuming and expensive.

% the weakly-supervised approaches are proposed without using any semantic attributes annotation.
The other approach without using semantic attributes labels is proposed.
Zhou \emph{et al.} \cite{ZhouL018} employed contrastive loss to cluster same view samples and infer multi-views features from single-view inputs.
Chu \emph{et al.} \cite{ChuSLLZW19} designed two viewpoint-aware metrics for similar viewpoints and different viewpoints. 
Furthermore, the unsupervised methods are proposed without orientation label. 
Jin \emph{et al.} \cite{0001LLPG21} model latent views from vehicle visual appearance directly by defining series of latent view clusters.
Bai \emph{et al.} \cite{BaiLLWD22} propose a novel 'Odd-One-Out' adversarial scheme to disentangle the orientation-invariant information.
Fortunately, the orientation information can be easier to predefined by prior knowledge or camera placements than semantic attributes labels.
Following \cite{BaiLLWD22}, in this paper, we devote to incorporating the orientation label information into the feature disentanglement process.
On the basis of this perspective, a simple but effective decoupling structure is proposed accompanied by two well-design pretext tasks.

\section{Dataset}
\subsection{Data collection and Annotation}
% drone Aerial
\textbf{Data collection}.
To solve the problem of data deficiency, we collect the first UAV cross-modality vehicle Re-ID benchmark, named UCM-VeID, which is mainly oriented towards urban transport.
The UAV platform is a DJI M300 RTK with a ZENMUSE H20T camera equipped with heterogeneous sensors, including RGB sensor and IR sensor.
The UAV employs cruise and spot rotating two kinds of sport modes to capture vehicles in five locations with complicated background, flying at altitudes ranging 60-100m.
A total number of 76 UAV videos are acquired by wide-angle camera and thermal camera, which contains 38 RGB UAV videos with the resolution $4056 \times 3040$ pixels per frame and IR UAV video with the resolution $640 \times 512$ pixels per frame, respectively.

\textbf{Annotation}.
The annotating of vehicle Re-ID, especially with multi modalities images, is the most time consuming step for construct a dataset.
Therefore, we devise a paradigm to make the annotation process efficient.
(1) First, we align RGB and IR videos by matching time and place, yielding a total of 38 matched video pairs for annotation.
In each pair videos, 2 frames per second are sampled to construct vehicle raw dataset.
(2) Owing to unavailable license plates in UAV videos and huge different appearance between RGB and IR modalities of same vehicle, experts can only manually locate and annotate matching vehicles based on auxiliary information such as time, location, targets' position and experience.
(3) Based on the above annotation information, original pictures dataset are sliced in batches to get slice dataset.
(4) In order to obtain valuable samples of slice target, we further invite volunteers to manually re-clean the results. 
(5) Finally, we labeled the acquired slice samples by camera number, ID number and image number to obtain the final dataset.

\begin{figure}[!t]                  
  \centering                    
  \includegraphics[width=\linewidth]{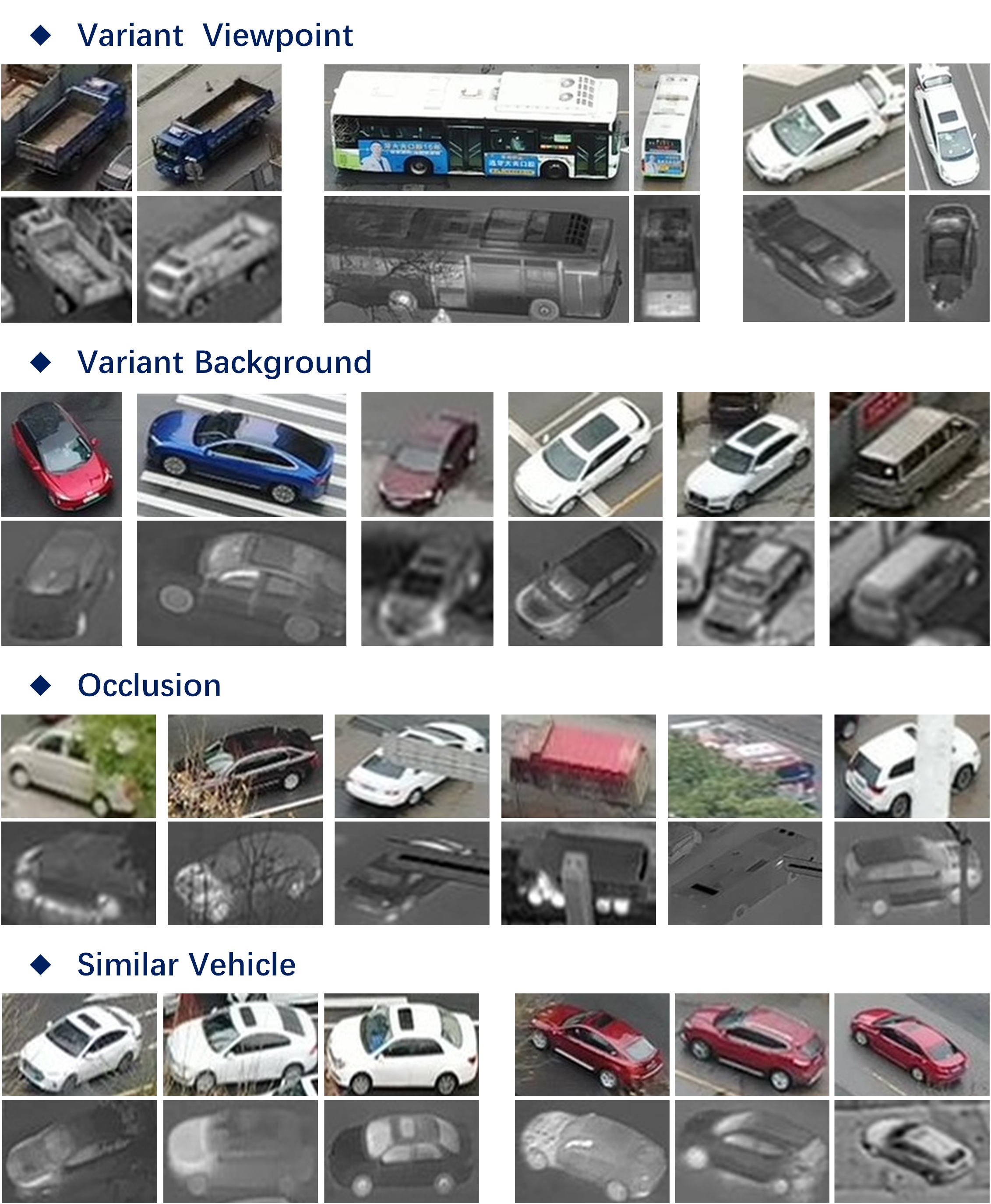}                  
  \caption{Challenge factors containing in UCM-VeID dataset. The first row of each challenge factor subgraph shows RGB samples, while the second displays IR samples. The images in every two columns are of the same vehicle.}
  \label{fig:C3}    
\end{figure}

\begin{table}[]
  % \small
  \centering   
  \scriptsize  
  % \footnotesize 
\begin{threeparttable} 
  \renewcommand{\arraystretch}{1.6}   
  \caption{Comparisons among other vehicle Re-ID datasets and UCM-VeID} 
  \label{tab:C1}
  \setlength{\tabcolsep}{0.6mm}{
  \begin{tabular}{lccccccc}
    \toprule[1pt] 
    Dataset              & VRAI          & UAV-VeID          & RGBN300          & MSVR310          & UCM-VeID                        \\ \hline
    Venue                & ICCV2019      & IJCV2021          & AAAI2020         & PR2023        & ours                               \\ \hline
    Images               & 137613        & 41917             & 50512            & 6216             & 29928                              \\ \hline
    Identities           & 13022         & 4601              & 300              & 310              & 753                                \\ \hline
    Modalities           & RGB           & RGB               & RGB+N            & RGB+N+T          & RGB/N                              \\ \hline
    Platform             & Mobile UAV    & Mobile UAV        & Fixed            & Fixed            & Mobile UAV                         \\ \hline
    Camera View          & Top-view      & Top-view          & Front-view       & Front-view       & Top-view                           \\ \hline
    Viewpoint            & Flexible      & Flexible          & Specific         & Specific         & Flexible                           \\ \hline
    Target State         & Motion        & Motion            & Stationary       & Stationary       & {Motion$\&$Stationary}             \\ \hline
    Vehicle Color        & 9             & No                & 9                & No               & 12                                 \\ \hline
    Vehicle Type         & 7             & No                & 8                & No               & 9                                  \\ \hline
    Orientation          & No            & 4                 & 8                & 8                & 8                                  \\ \hline
    Occlusion            & No            & No                & Yes              & No               & Yes                                \\   
    \bottomrule[1pt]
  \end{tabular}
  \begin{tablenotes}
    \item[*] The '$+$' represents multi modality and '/' represents cross modality.
  \end{tablenotes}}
\end{threeparttable} 
\end{table}
\subsection{Dataset Description}
\begin{figure}[!t]
  \centering
  \subfigure[]{\label{fig:c4_a}\includegraphics[width=0.48\linewidth]{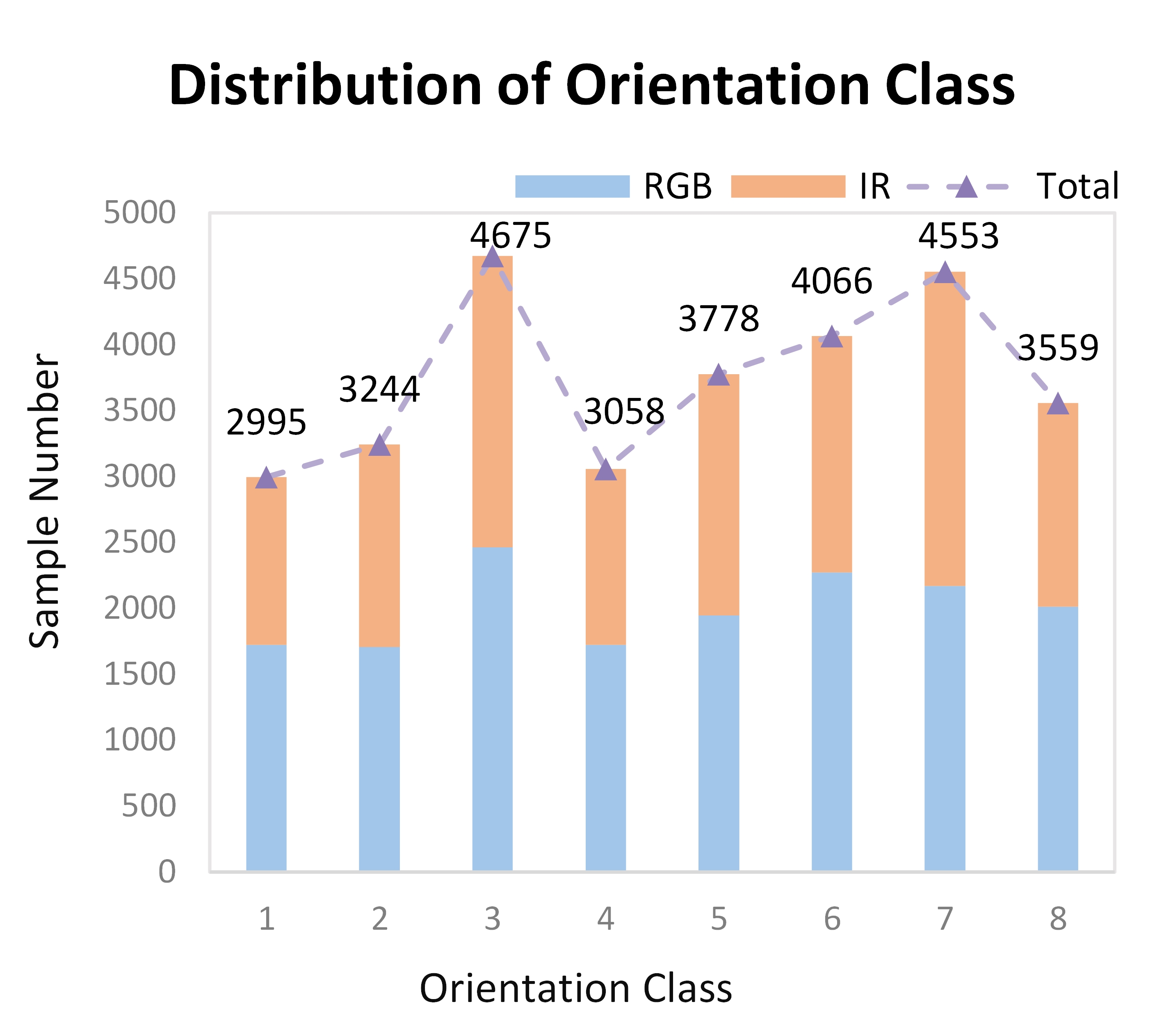}}
  \subfigure[]{\label{fig:c4_b}\includegraphics[width=0.48\linewidth]{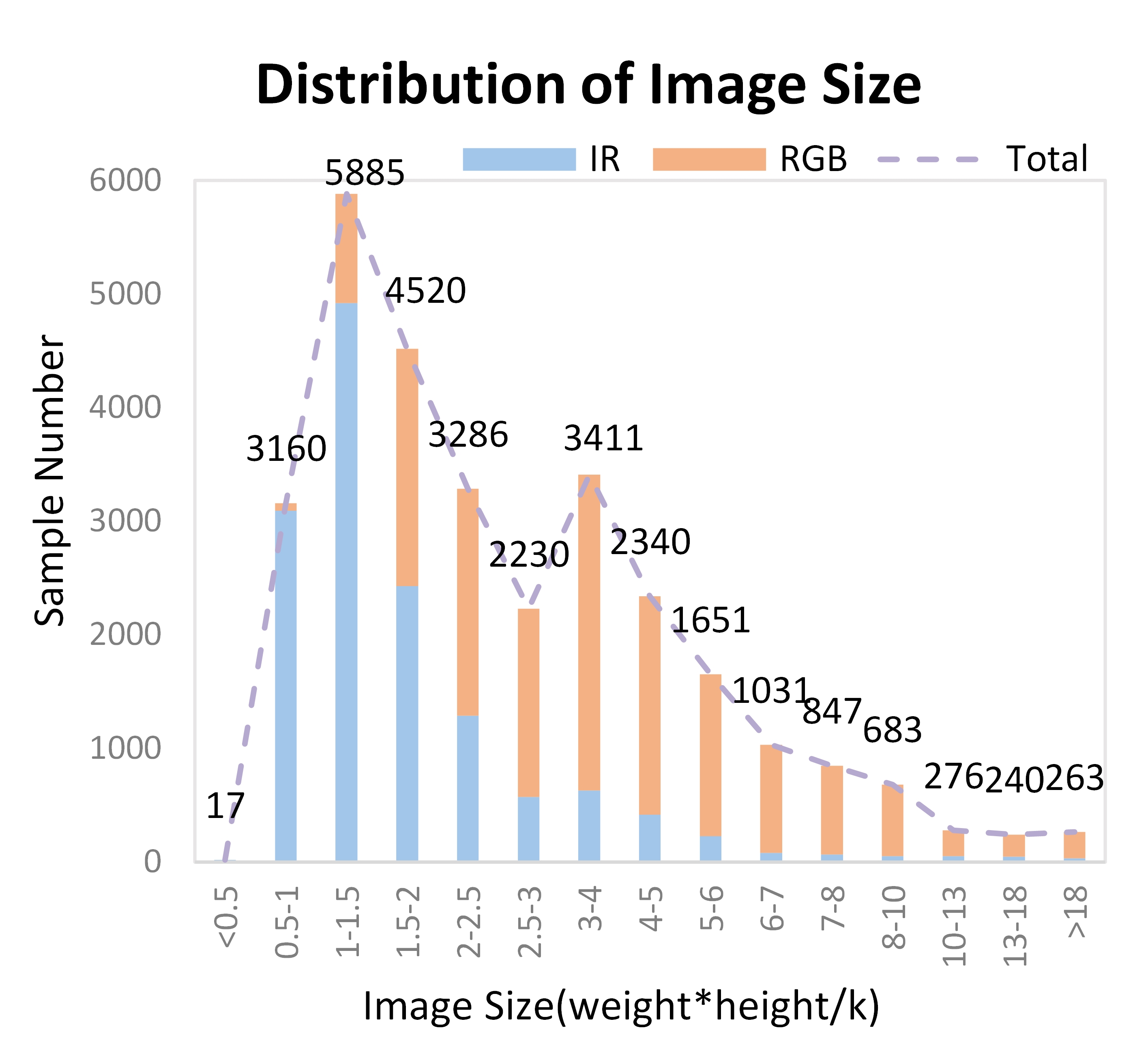}}
  \subfigure[]{\label{fig:c4_c}\includegraphics[width=0.48\linewidth]{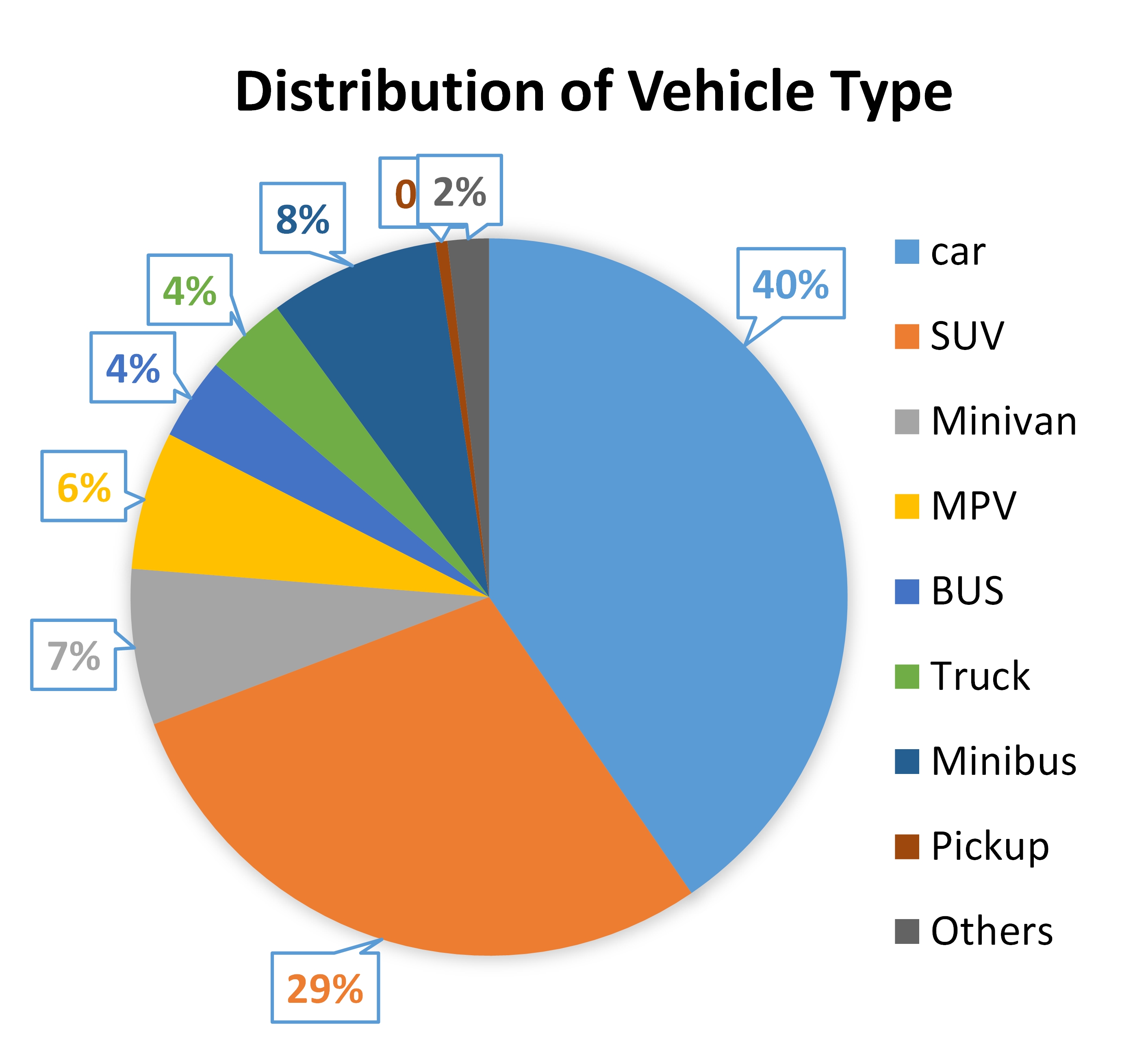}}
  \subfigure[]{\label{fig:c4_d}\includegraphics[width=0.48\linewidth]{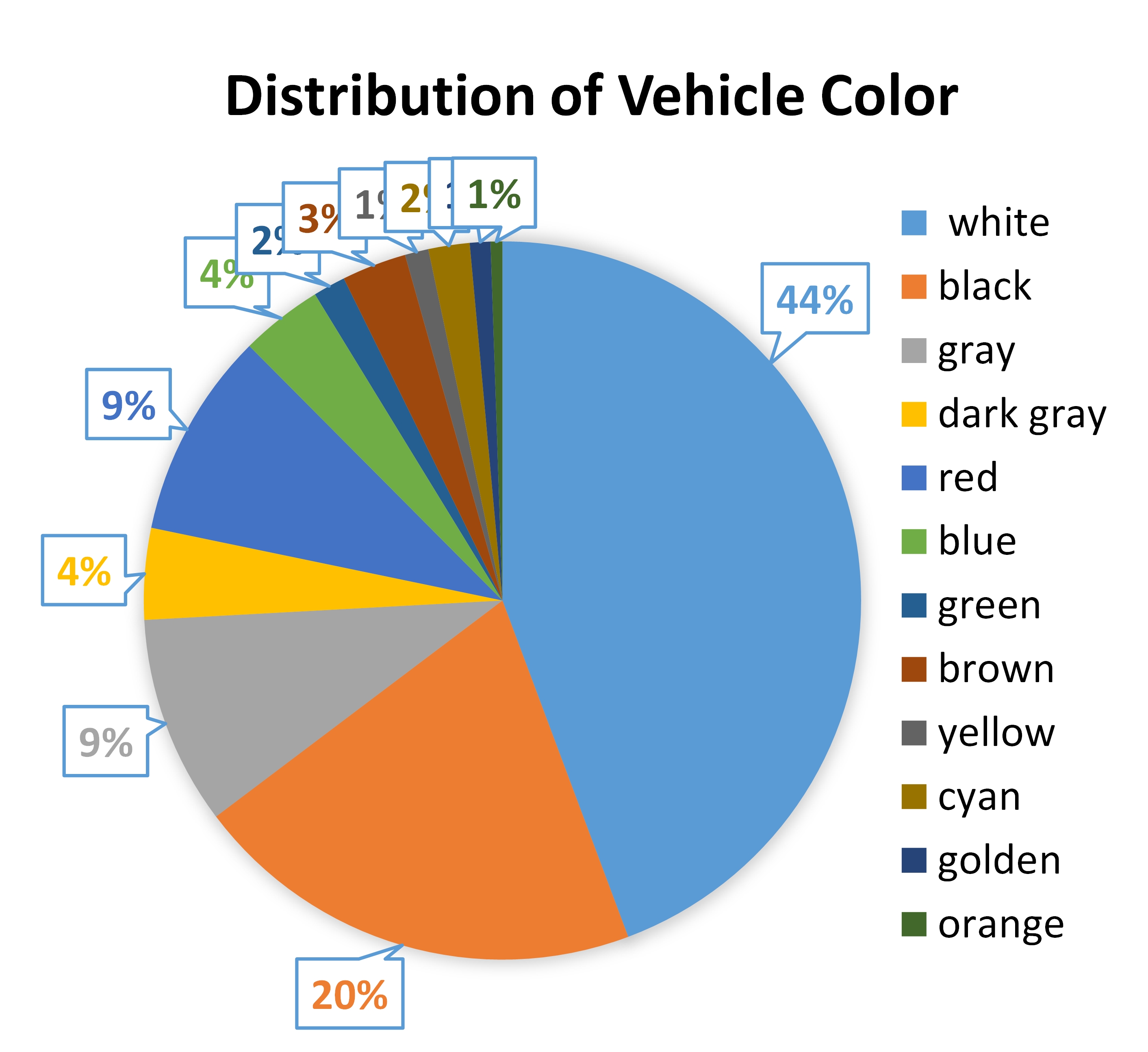}}
  \subfigure[]{\label{fig:c4_e}\includegraphics[width=0.48\linewidth]{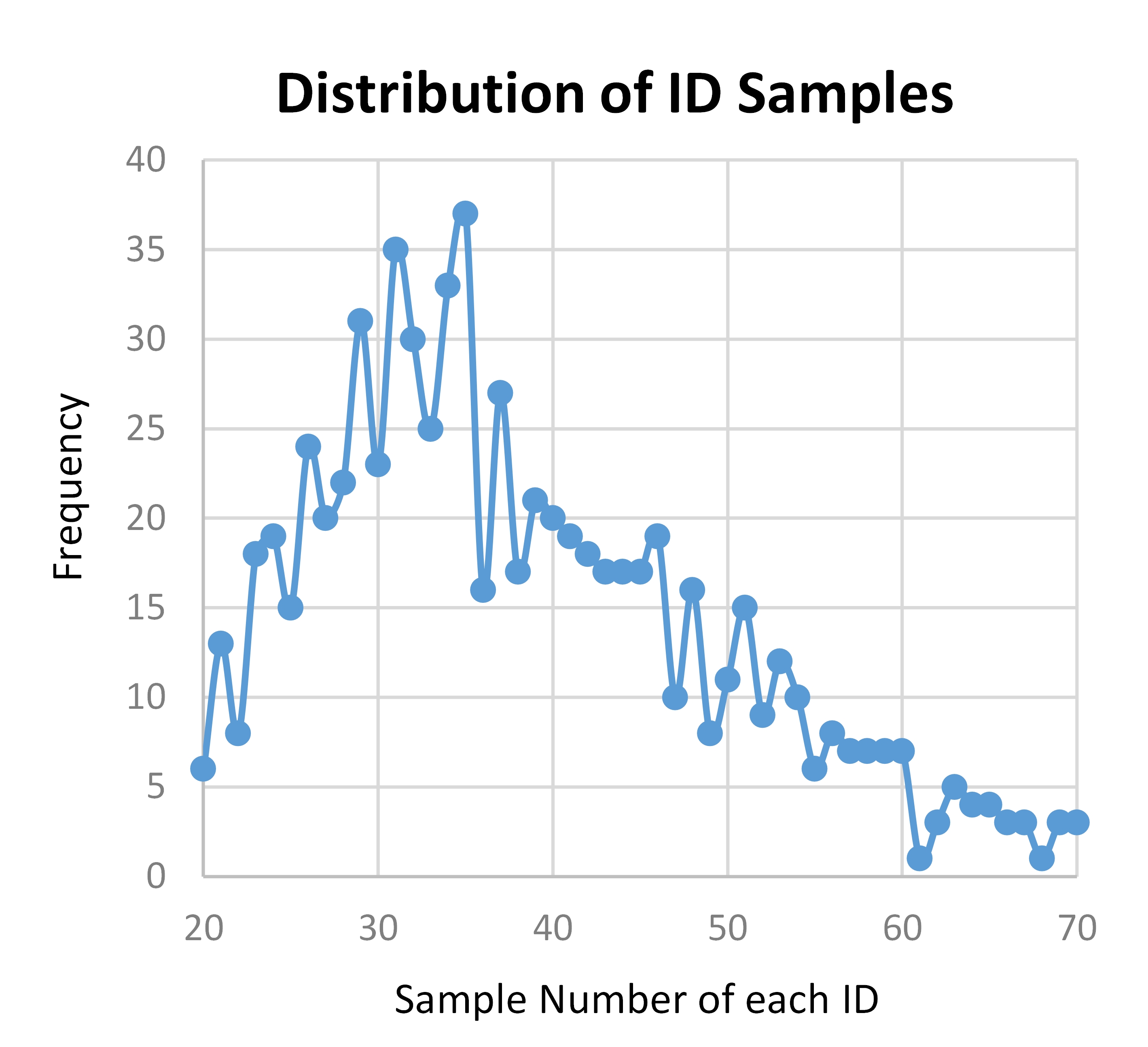}}
  \subfigure[]{\label{fig:c4_f}\includegraphics[width=0.48\linewidth]{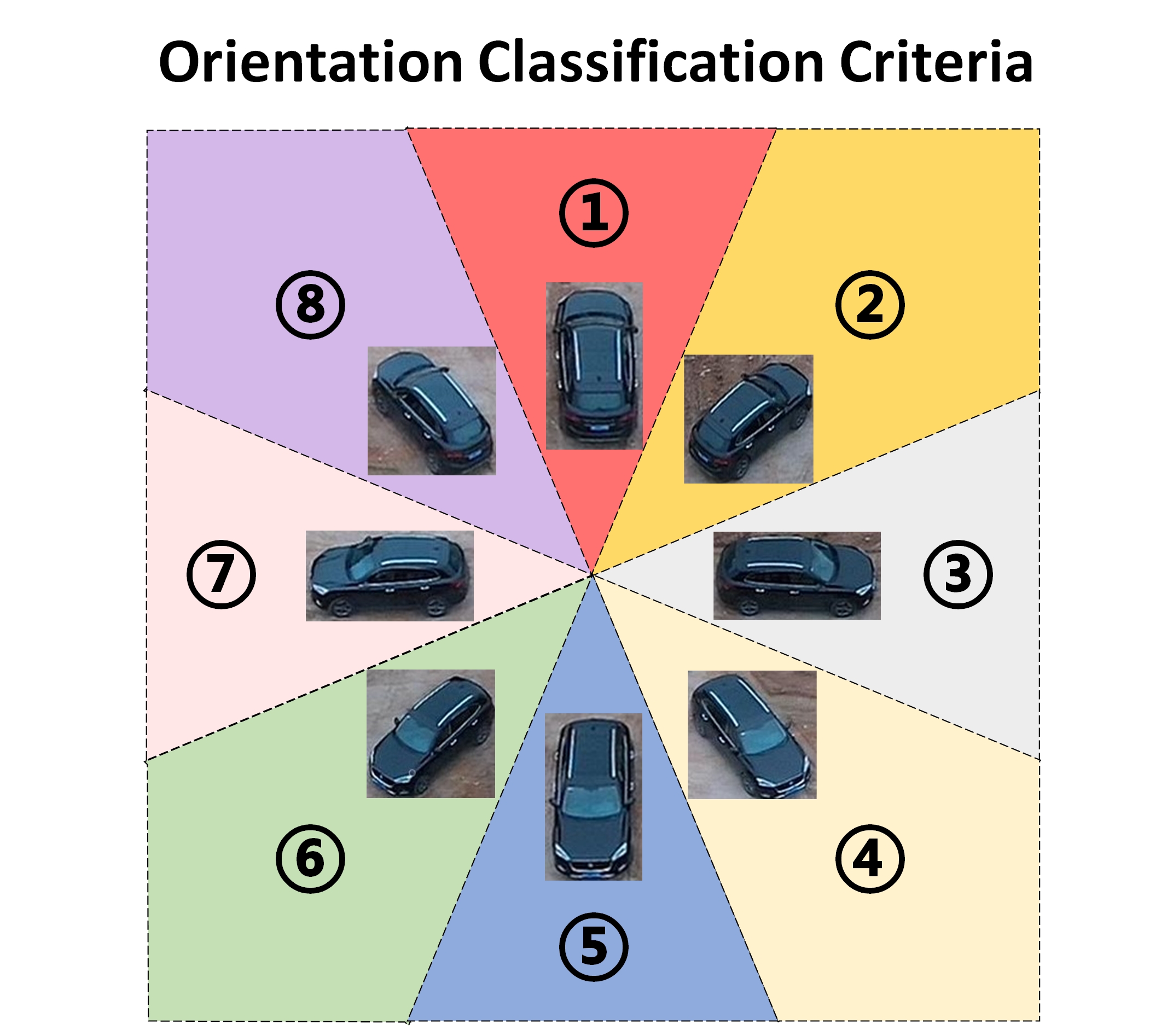}}
  %\vspace{-2ex}
  \caption{The statistics of UCM-VeID dataset, including the distributions of the (a) orientation class; (b) image size; (c) vehicle type; (d) vehicle color; (e) ID samples and (f) orientation classification criteria.}
  \label{fig:C4}
\end{figure}

We achieve 753 vehicle identities in UCM-VeID, where 602 identities with 23555 images used as training set and 151 identities with 4907 images for testing. 
Each vehicle has at least 16 samples including 8 RGB samples and 8 IR samples.
Some examples of UCM-VeID are shown in the Fig.\ref{fig:C3}.
For further research, we completed the annotation of the vehicle's view attributes by referring to the vehicle orientation and obtained a total number of 29928 view annotations with 8 categories, as illustrated in Fig.\ref{fig:C4} (a). 
The criteria for classifying vehicle orientation categories are shown in Fig.\ref{fig:C4} (f).
What's more, we calculated the distribution of data characteristics in UCM-VeID, such as image size, vehicle type, vehicle color, and ID samples, as visualized in Fig.\ref{fig:C4} (b)-(e).
Comparing with other vehicle Re-ID dataset referring as Table \ref{tab:C1}, our dataset has the distinctive features as following:

\begin{enumerate}[]
  \item \textbf{Multi modal information.} Compared with the existing RGB-based datasets, UCM-VeID contains RGB and IR two different modal data to support cross-modality vehicle Re-ID task, which is a potential application to realize full-time vehicle Re-ID.
  \item \textbf{Flexible viewpoint, scale and orientations.} Our dataset is collected by mobile camera fixed on UAV platform. Owing to the mobility and flexibility of UAV, the vehicle images in UCM-VeID presents various viewpoints, scales and orientations, coupled with the influence of different modal data, making cross-modality vehicle Re-ID a more challenging task.
  \item \textbf{Rich attribute annotations.} In order to exploit the full potential of UCM-VeID, rich attributes are annotated for providing more auxiliary information, including 12 vehicle colors, 9 vehicle types and 8 class vehicle orientations. 
  \item \textbf{More realistic task.} Different from multi-spectral vehicle Re-ID dataset, UCM-VeID is collected by UAV platform in top view. The collection mode simple, fast and unconstrained and the images obtained do not require manual alignment. As a result, UCM-VeID is closer to the realistic world and more widely used. 
  \end{enumerate}

\section{Hybrid Weights Decoupling Network} \label{sec:4}
In this section, we propose a hybrid weights decoupling network (HWDNet) for cross-modality vehicle Re-ID.
We model the two issues of cross-modality vehicle Re-ID introduced in Section \ref{sec:1} and describe the pipeline of HWDNet.
In Section \ref{sec4:2}, we introduce a novel siamese network with a weight restrainer to learn the shared representation of shallow layers.
Besides, three decoupling strategies are presented in Section \ref{sec4:3} to split features into two unrelated parts guided by two pretext tasks.
The description of HWDNet is detailed as following.

\subsection{Overview}
% We describe the cross-modality vehicle Re-ID problem and briefly introduce the pipeline of our method.

\begin{figure*}[!t]                  
  \centering                    
  \includegraphics[width=\linewidth]{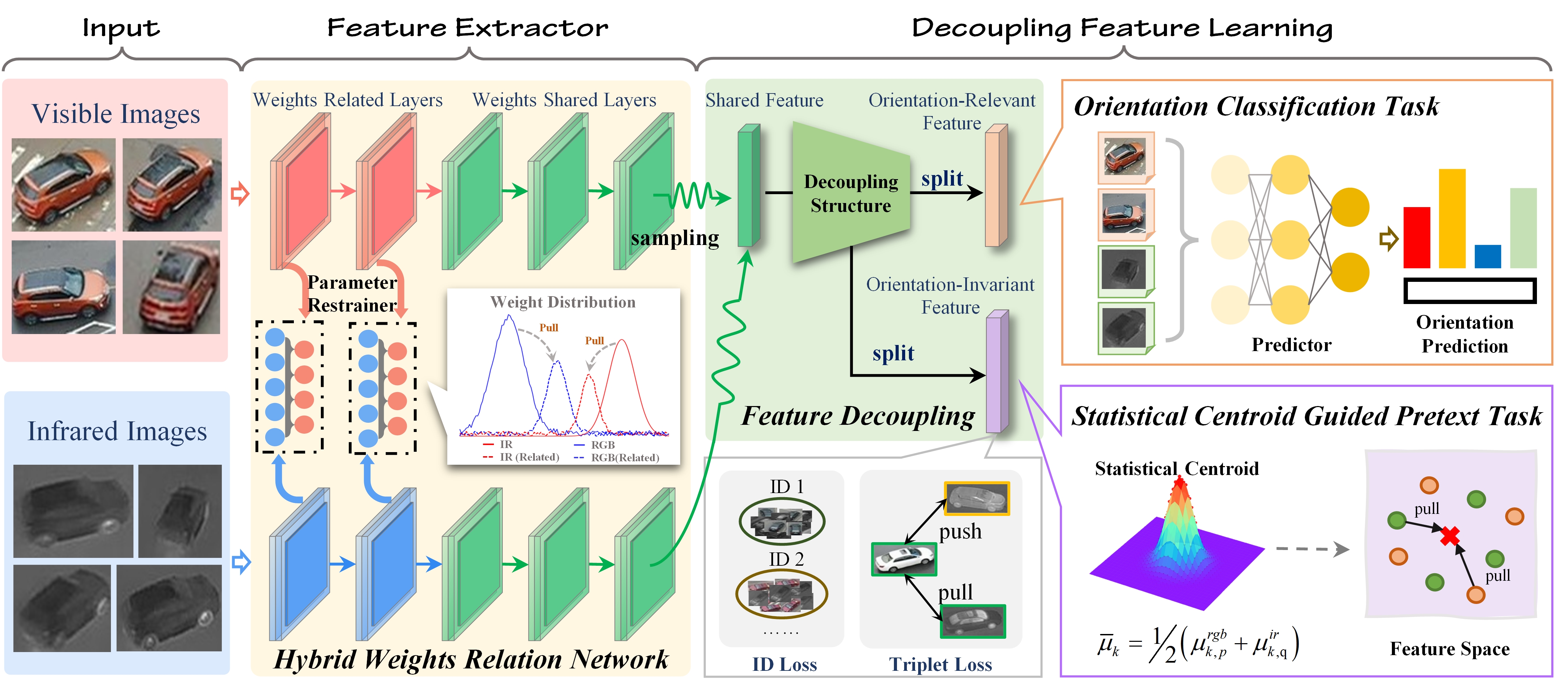}
  % \caption{The framework of cross-modality orientation invariant feature learning (CM-OIFL). }      
  \caption{The framework of our proposed Hybrid Weights Decoupling Network (HWDNet). It mainly contains two components: a hybrid weights siamese network and a simple but effect decoupling structure. The hybrid weights siamese network can extract shared features from two relations network layers, deep shared layers by sharing parameters and shallow specific layer with Parameter Related Layers. The decoupling structure divide the feature into two parts: orientation-relevant feature and orientation-invariant feature directed by orientation classification task and statistical centroid guided pretext task, respectively.}              
  \label{fig:C5}    
\end{figure*}

\textbf{Problem definition}.
Generally, UCM-VeID dataset $\mathcal{D}$ is composed of a training set $\mathcal{D}_\emph{tr}$ and a testing set $\mathcal{D}_\emph{te}$. 
$\mathcal{D}_\emph{te}$ would be further divided into query and gallery sets, where they contains different modal samples.
Supposing that, a mini-batch of $\mathcal{D}_\emph{tr}$ consists of $M$ RGB images $\mathcal{X}_{rgb}=\{x_i^{rgb}\}_{i=1}^{M}$ and $N$ IR images $\mathcal{X}_{ir}=\{x_i^{ir}\}_{i=1}^{N}$ with corresponding identities labels $\mathcal{Y}=\{y_i\}_{i=1}^{M}$ and $\mathcal{Y}=\{y_i\}_{i=1}^{N}$ respectively.
% where $K$ stands for the numbers of the ID classes.
In order to complete the cross-modality vehicle Re-ID task, the learning object is to learn a feature extractor $f(\cdot)$, which is able to transform $x_i^{rgb}$ and $x_i^{ir}$ to feature vectors $z_i^{rgb}$ and $z_i^{ir}$ as
\begin{equation}
  \begin{gathered} 
    z_i^{rgb} = f(x_i^{rgb})\\ 
    z_i^{ir} = f(x_i^{ir})
  \end{gathered},
\end{equation}
where $z_i^{rgb}$ and $z_i^{ir}$ have closest distance in Euclidean space, if $x_i^{rgb}$ and $x_i^{ir}$ belong to the same ID.
%  with different orientations.
% During the testing phase, given a query sample, the samples of same ID in gallery are obtained without using orientation-related features.

\textbf{Overall Model Structure}.
The overall architecture of our proposed HWDNet is illustrated in Fig.\ref{fig:C5}.
The same number of RGB samples and IR samples with same ID are selected as the inputs of feature extractor for feature learning.
The feature extractor is a two-stream network, which could be divided into two parts according to the relation of the weights: weights related layers and weights shared layers.
In weights related layers, the weights of RGB branch and IR branch are related to each other, which can be achieved by the weight restrainer with linear loss function we designed.
While, weights shared layers, RGB branch network and IR branch network utilize the same architecture and parameters to extract shared representations.
Then, the obtained shared features is divided to two unrelated features.
Accordingly, we design orientation classification task and statistical centroid guided pretext task to guide the two parts of features to learn orientation-relevant and orientation-invariant characteristics respectively.
Moreover, we use ID loss and triple loss to make the orientation-invariant feature more discriminative.

\subsection{Feature extractor}\label{sec4:2}
To eliminate the cross-modality semantic gap while keeping the discriminant of representations, the existing two-stream network based methods typically share the weights of deep layers while make the weights independent in shallow layers.
As a result, the shallow layers aim to learn modality-specific information through the same network structure with different weights, and deep layers use the same network parameters of two modality branch networks for excavating modality-shared features.
However, the shallow layers are unable to eliminate the cross-modality discrepancy and even lose some shared features. 

To deal with this obstacle, the core perspective of this work is making the weights of shallow layers related. 
In detail, we allow the weights of shallow layers in the two-stream network to be different but prevent them from being too far with the help of a devised weight restrainer.
This weight restrainer is realized by a fully connected layer in order to optimize a specific linear transform objective function.
\par 
\begin{figure*}[!t]                  
  \centering                    
  \includegraphics[width=\linewidth]{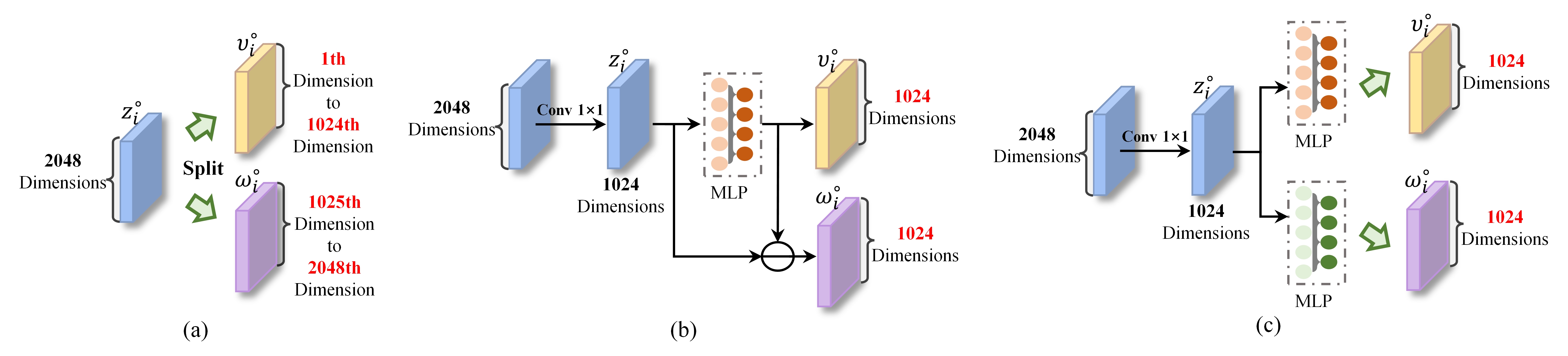}                  
  \caption{Feature Decoupling Structure. (a) Feature split structure; (b) Feature subtraction structure and (c) Feature prediction structure}                  
  \label{fig:C6}    
\end{figure*}
\textbf{Weight restrainer.} Mathematically, the weights of $i$-th layer in feature extractor for RGB (IR) modality denote as $\boldsymbol{W}_{i}^{RGB} (\boldsymbol{W}_{i}^{IR})$. 
Then, to establish the relation between the weights of $i$-th layers in the two-stream network, the weight restrainer exploits several fully connected layers, whose weights and bias refer to $\boldsymbol{A}=\{\boldsymbol{a}_{1}, \boldsymbol{a}_{2},\cdots,\boldsymbol{a}_{5}\}$ and $\boldsymbol{B}=\{\boldsymbol{b}_{1}, \boldsymbol{b}_{2},\cdots,\boldsymbol{b}_{5}\}$, to transform the weight of $i$-th layer in RGB-stream network as follows:
\begin{equation}
  \hat{\boldsymbol{W}}_{i}^{RGB}=\boldsymbol{a}_{i} \cdot \boldsymbol{W}_{i}^{RGB} + \boldsymbol{b}_{i}, i=1,2,\cdots,5.
\end{equation}
\par 
Then, the distance between transformed RGB weight and IR weight is formulated as:
\begin{equation}
  D(\hat{\boldsymbol{W}}_{i}^{RGB}, \boldsymbol{W}_{i}^{IR}) = ||\hat{\boldsymbol{W}}_{i}^{RGB} - \boldsymbol{W}_{i}^{IR}||_{2}
\end{equation}
where $||\cdot||_{2}$ is the $l_2$-norm of a matrix. 
For the sake of establishing the relation between the weights of $i$-th layers in the two-stream network, we are going to optimize the following objective function: 
\begin{equation}
  \mathcal{L}_{wr} = \sum_{n = 1}^{5} \alpha_{i} \cdot ||\hat{\boldsymbol{W}}_{i}^{RGB} - \boldsymbol{W}_{i}^{IR}||_{2}
\end{equation}
and 
\begin{equation}     
  \alpha_{i} =      
  \begin{cases}         
    1, & \text{if the i-th layers are related} \\                 
    0, & \text{if the i-th layers are not related}      
  \end{cases} 
\end{equation}
\par
\textbf{Identity classification supervised task.} 
The existing cross-modality Re-ID methods usually adopt identity classification supervised task with identity loss $\mathcal{L}_{ID}$ to learn the discriminative feature representations.
% Meanwhile, a self-supervised contrastive learning task with triple loss $\mathcal{L}_{tri}$ is employed to further learn fine-grained discriminant features based on the relationship among different identities across two modalities.
% In this paper, we utilize the identity loss $\mathcal{L}_{ID}$ and triple loss $\mathcal{L}_{tri}$ as the baseline loss function of the proposed method, which is presented as 
% \begin{equation}
%   \mathcal{L}_{base} = \mathcal{L}_{ID} + \mathcal{L}_{tri},
% \end{equation}
Typically, we take the network output ${z_i^{\circ }}$ through a $softmax$ operation to get the sample's identity prediction $p_i^{\circ }=\frac{exp({z_i^{\circ }})}{\sum_{k = 1}^{K}{exp({z_i^{\circ }})}} $, where $\circ=\{rgb,ir\}$, $K$ is the total number of identities.
And then cross entropy loss is adopted to obtain identity loss $\mathcal{L}_{ID}$ shown as 
\begin{equation}
  \mathcal{L}_{ID} = -\sum_{i = 1}^{M}{y_i}\log{p_i^{rgb}}-\sum_{i = 1}^{N}{y_i}\log{p_i^{ir}}.
  \label{eq:3}
\end{equation}
\par
\textbf{Identity discrimination self-supervised task.} 
A self-supervised contrastive learning task with triple loss $\mathcal{L}_{tri}$ is employed to further learn fine-grained discriminant features based on the relationship among different identities across two modalities.
The cross-modality triple loss purposes to shrink the feature distance between the anchor and positive sample and enlarge the feature distance between the anchor and positive sample, which is represented by
\begin{equation}
\begin{split}
  \mathcal{L}_{tri} = 
  &\sum_{n = 1}^{N}[\rho+\min d(z_i^{rgb},z_j^{ir})-\max d(z_i^{rgb},z_k^{ir})]_+ + \\
  &\sum_{n = 1}^{N}[\rho+\min d(z_i^{ir},z_j^{rgb})-\max d(z_i^{ir},z_k^{rgb})]_+
\end{split}
\end{equation}
where $[]_+ = \max(\cdot ,0)$, $\rho$ represents the margin parameter and $d(\cdot)$ is Euclidean space feature distance calculation paradigm.
\par 
To sum up, the overall objective function of the feature extractor is modeled as follows:
\begin{equation} 
  \mathcal{L}_{extractor} = \mathcal{L}_{wr} + \mathcal{L}_{ID} + \mathcal{L}_{tri}. 
\end{equation}

\subsection{Orientation-Invariant Feature Separation}\label{sec4:3}
For eliminating the influence of vehicle orientation on the UAV vehicle Re-ID task, we investigate the feature decoupling strategy and devise a orientation classification task to separate orientation-relevant features from the latent space.

\textbf{Feature decoupling strategy.} 
As demonstrated in Fig. \ref{fig:C6}, three feature decoupling structures are presented in this work. 
They are named feature split structure, feature subtraction structure and feature prediction structure respectively.
\par
\emph{(1) Feature split structure: }Feature split structure refers to the most general decoupling structure, which rudely splits the latent representation into two parts along the feature dimension.
Then, a split sub-feature is utilized as the orientation-relevant feature $\upsilon_i^{\circ}$ and the rest one are regraded as the orientation-invariant feature $\mu_i^{\circ}$ .
\par 
\emph{(2) Feature subtraction structure: }The theoretical basis of the feature split structure is that, each part of decoupling results is independent to each other.
However, this theoretical basis does not always hold in practice, leading to semantic information loss in the splitting features.
For the sake of addressing above issue, as illustrated in Fig. \ref{fig:C6} (b), a novel decoupling structure is proposed based on a prediction strategy.
To be more specific, given a latent representation ${z_i^{\circ}}$, a multilayer perceptron (MLP) $\mathcal{G}(\cdot)$ is utilized as a predictor to generate the orientation-relevant feature $\upsilon_i^{\circ}$ by
\begin{equation}
    \upsilon_i^{\circ} = \mathcal{G}({z_i^{\circ}}).
\end{equation}
After that, considering the complementary characteristic of decoupled features, the orientation-invariant feature $\mu_i^{\circ}$ can be obtained approximately by the minus operation as
\begin{equation}
  \mu_i^{\circ} = {z_i^{\circ}} - \upsilon_i^{\circ}.
  \label{eq:6}
\end{equation}
\par  
\emph{(3) Feature prediction structure: }Compared with the feature split structure, the feature prediction structure generate the decoupled features only based on the original latent feature without considering the relationship between the decoupled features.
The intuition of this decoupling structure is that using the relationship between the decoupled features as the prior information for decoupling structure design may have the adverse impact. 
To this end, two multilayer perceptron (MLP) $\mathcal{G}_{r}(\cdot)$ and $\mathcal{G}_{u}(\cdot)$ are utilized as the predictors to generated the corresponding decoupled features:
\begin{equation}     
  \upsilon_i^{\circ} = \mathcal{G}_{r}({z_i^{\circ}}). \\
  % \mu_i^{\circ} = \mathcal{G}_{u}({z_i^{\circ}}). 
\end{equation}
\begin{equation}        
\mu_i^{\circ} = \mathcal{G}_{u}({z_i^{\circ}}).  
\end{equation}
\par
\textbf{Orientation Classification Task.}
Decoupling structure can not restrict the semantic attributes of decoupled features.
To this end, following \cite{Decoupling2019}, we conduct a orientation classification task to make full use of the orientation annotations of UCM-VeID dataset.
Let us denote the corresponding orientation labels of the orientation-relevant feature $\upsilon_i^{\circ}$ as $r_i$.
The intention of orientation classification task is to establish mapping relationship between orientation-relevant feature $\upsilon_i^{\circ}$ and orientation label $r_i^{\circ}$ by a orientation classifier $\mathcal{C}(\cdot)$, which can be modeled as $\hat{r_i^{\circ}} \triangleq  \mathcal{C}(\upsilon_i^{\circ})$.
Analogous to Eq. \ref{eq:3}, the orientation classifier can be fitted by the following loss function
\begin{equation}
  \mathcal{L}_R = -\sum_{i = 1}^{M}{r_i^{rgb}}\log{\hat{r}_i^{rgb}}-\sum_{i = 1}^{N}{r_i^{ir}}\log{\hat{r}_i^{ir}},
\end{equation}
where the first term represents the orientation classification loss of RGB samples, while the second term stands for the one of IR samples.

\subsection{Orientation-Invariant Feature Refining}
% To address above issues, we design two subtasks based on the orientation physics prior, namely unbalanced orientation classification and cross-modal extraction, respectively.
% One of which is used to extract orientation related features and the other is used to learn orientation invariant features. 
% The details are shown below.
The aforementioned orientation classification task is inadequate to make orientation-relevant features perfectly describe all the orientation information due to the discrete orientations.
Therefore, orientation-invariant features derived by Eq. \ref{eq:6} may contain a little orientation information, which has negative impact on Re-ID task. 
To achieve a better Re-ID performance, we further refine the orientation-invariant features with feature similarity enforcement and cross-modality statistical centroid construction.

\textbf{Feature Similarity Enforcement.}
% Motivated by [], feature average is one of the most commonly used strategy to calculate the orientation-unrelated feature.
% The orientation-unrelated feature is not available in testing stage, but can be used as a reference to generate orientation-invariant feature
Motivated by \cite{Decoupling2019}, an effective refining method is to enforce similarity between orientation-invariant features of the same ID with different orientations.
Feature average strategy can be used to calculate the reference information for feature similarity enforcement. 
In the other word, we could take the calculated average feature as statistical centroid to figure out the features similarity.
Formally, the orientation-invariant feature of $j$-th sample in $k$-th ID is denoted as $\mu_{k,j}^{\circ}$, and the process of similarity enforcement can be modeled as 
\begin{equation}
  \mathcal{L}_C = \sum_{k = 1}^{K} \sum_{p = 1}^{P}s(\mu_{k,p}^{rgb}, \bar{\mu}_{k}^{rgb}) + 
        \sum_{k = 1}^{K} \sum_{q = 1}^{Q}s(\mu_{k,q}^{ir}, \bar{\mu}_{k}^{ir}) 
  \label{eq:8}
\end{equation}
where statistical centroids of RGB and IR modalities are
\begin{equation}
  \bar{\mu}_{k}^{rgb} = \frac{1}{P} \sum_{p = 1}^{P}\mu_{k,p}^{rgb},
\end{equation}
\begin{equation}
  \bar{\mu}_{k}^{ir} = \frac{1}{Q} \sum_{q = 1}^{Q}\mu_{k,q}^{ir},
\end{equation}

% Considering the characteristic variability and unit dimension inconsistences of different features, we adopt the Mahalanobis Distance \cite{Mohammed2018} for feature similarity measurement as
% \begin{equation}
%   s(\mu_{k,j}^{\circ}, \bar{\mu}_{k}^{\circ}) = \sqrt{(\mu_{k,j}^{\circ}-\bar{\mu}_{k}^{\circ})^{T}(\Sigma_{k}^{\circ})^{-1}(\mu_{k,j}^{\circ}-\bar{\mu}_{k}^{\circ})}, 
%   \label{eq:11}
% \end{equation}
% where $\Sigma_{k}^{\circ}$ refers to the covariance matrix of RGB or IR samples for the $k$-th ID.

\textbf{Cross-Modality Statistical Centroid Construction}. 
\begin{figure}[!t]                  
  \centering                    
  \includegraphics[width=\linewidth]{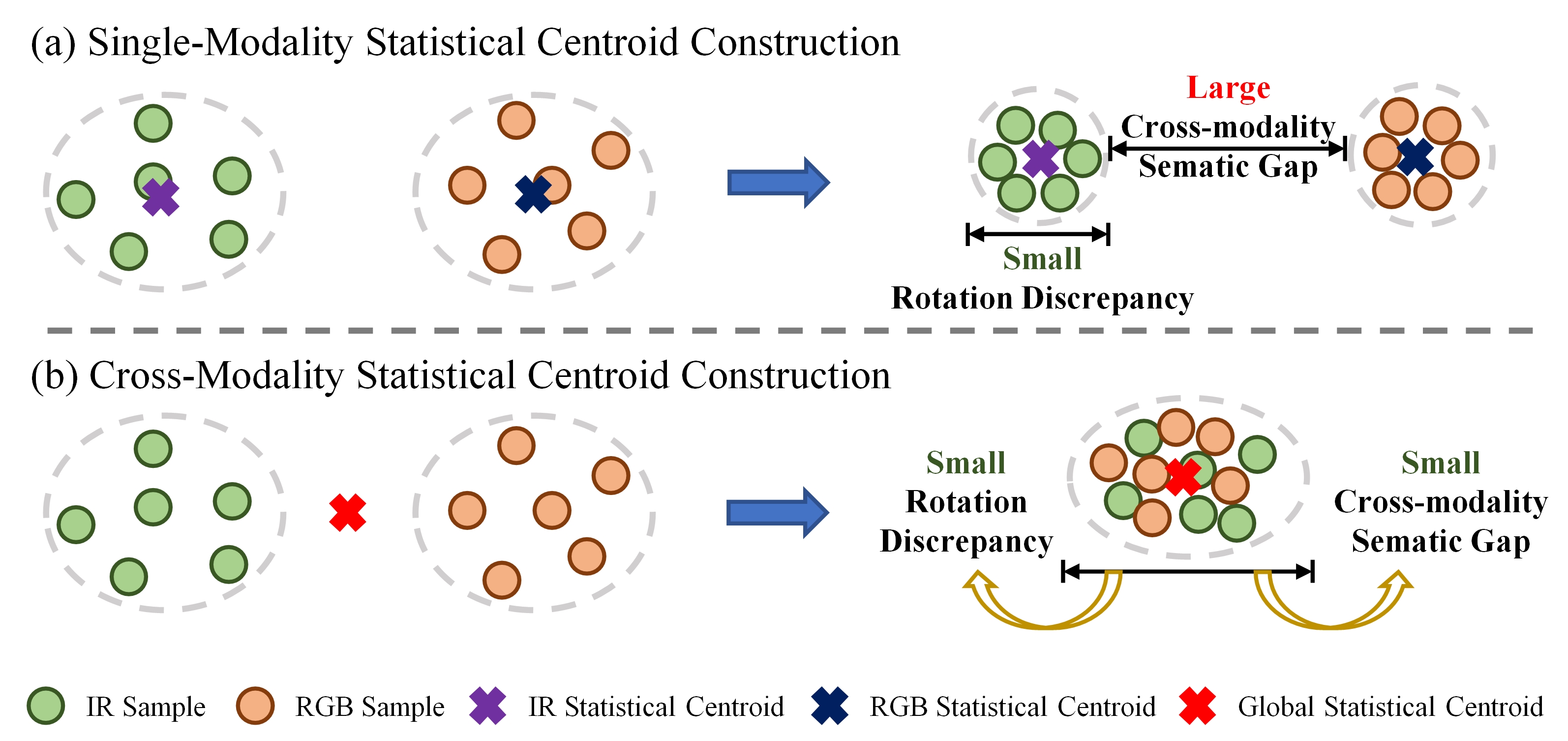}                  
  \caption{Statistical Centroid Construction. (a) Single-Modality; (b) Cross-Modality.}                 
  \label{fig:C7}    
\end{figure}
As shown in Fig. \ref{fig:C7} (a), if the statistical centroid is constructed on the basis of single modality data, we could achieve a smaller orientation discrepancy but a larger cross-modality semantic gap.
To decrease the cross-modality semantic gap and the orientation discrepancy simultaneously, we come up with a cross-modality statistical centroid construction method by using the global statistical information of all the training samples belonging to same ID.
The illustration of the proposed method is established in Fig. \ref{fig:C7} (b).
Mathematically, the mean vector $\bar{\mu}_{k}$ of global statistical centroid for $k$-th ID can be presented as
\begin{equation}
  \bar{\mu}_{k} = \frac{1}{2} \left({\frac{1}{P} \sum_{p = 1}^{P}\mu_{k,p}^{rgb} + \frac{1}{Q} \sum_{q = 1}^{Q}\mu_{k,q}^{ir}}\right).
\end{equation}

% In a similar manner, the covariance matrix of global statistical centroid $\Sigma_{k}$ for $k$-th ID is described as
% \begin{equation}
%   \Sigma_{k}^2 = (\Sigma_{k}^{rgb})^2 + (\Sigma_{k}^{ir})^2 + 2\Sigma_{k}^{rgb, ir},
% \end{equation}
% where $\Sigma_{k}^{rgb, ir}$ refers to the joint covariance matrix.
According to Eq. \ref{eq:8}, we can reformulate the similarity measurement $s(\cdot, \cdot)$ and similarity enforcement loss $\mathcal{L}_C$ as
\begin{equation}
  \mathcal{L}^{'}_C = \sum_{k = 1}^{K} \sum_{p = 1}^{P}s(\mu_{k,p}^{rgb}, \bar{\mu}_{k}) + 
        \sum_{k = 1}^{K} \sum_{q = 1}^{Q}s(\mu_{k,q}^{ir}, \bar{\mu}_{k}).
\end{equation}

In summary, the total loss function of the method can be expressed as
\begin{equation}
  \mathcal{L}_{total} = \mathcal{L}_{wr} + \mathcal{L}_{ID} + \mathcal{L}_{tri} +\mathcal{L}_R + \mathcal{L}^{'}_C 
\end{equation}

\begin{table*}[!t]   
  \centering   
  % \scriptsize  
  % \footnotesize 
  \small
  \renewcommand{\arraystretch}{1.35}   
  \begin{threeparttable}   
    \caption{Comparisons between the proposed HWDNet and some state-of-the-art methods on UCM-VeID datasets. The rank-k (\%) and mAP (\%) are demonstrated.} 
    \label{tab:C2}
    \tabcolsep=5pt   
    %\begin{minipage}{20cm}   
    \setlength{\tabcolsep}{2.5mm}{   
      \begin{tabular}{c|ccc|ccc|ccc|ccc}   
        \toprule[1pt] %\toprule define the top line width   
        \hline
        \multirow{3}{*}{Method}  & \multicolumn{6}{c}{Single-shot}\vline &\multicolumn{6}{c}{Multi-shot}\\
        \cline{2-13}
        &\multicolumn{3}{c}{IR2RGB} \vline   & \multicolumn{3}{c}{RGB2IR} \vline & \multicolumn{3}{c}{IR2RGB} \vline& \multicolumn{3}{c}{RGB2IR}\\
        \cline{2-13}
        &rank1  & rank10  & mAP &rank1  & rank10  & mAP &rank1  & rank10  & mAP  &rank1  & rank10  & mAP \\      
        \hline
        AGW\cite{AGW2020} &  28.36& 72.84& 42.74& 27.13& 73.13& 42.06& 34.87& 81.65& 30.17& 34.10& 82.14& 29.27 \\ \hline
        DDAG\cite{DDAG2020} &  25.41& 69.99& 39.79& 25.05& 70.59& 39.92& 34.49& 80.38& 28.53& 33.80& 79.94& 27.55 \\ \hline
        CGRNet\cite{Y.Feng2021} &  27.93& 73.64& 42.99& 28.77& 73.10& 43.38& 35.04& 82.87& 30.64& 36.96& 83.27& 30.41 \\ \hline
        MMD-ReID\cite{Jambigi2021} & 28.98& 72.85& 43.39& 28.52& 73.38& 43.09& 40.34& 86.09& 31.10& 39.04 & 85.85& 30.30 \\ \hline
        DGTL \cite{H.Liu2021}  & 29.26& 72.70& 43.61& 27.59& 71.94& 42.14& 35.06& 80.65& 30.68& 37.05& 81.35& 30.04 \\ \hline
        LbA\cite{H.Park2021} & 27.22& 78.22& 43.52& 27.92& 78.50& 43.99& 34.78& 86.07& 29.71& 36.63& 83.37 & 30.43 \\ \hline
        Baseline &  26.24& 75.01& 42.19& 28.17& 77.39& 43.66& 32.59 & 82.21& 30.18& 34.18& 84.28& 31.09 \\ \hline
        HWDNet &  \textbf{31.57} & \textbf{78.85} & \textbf{46.09} & \textbf{30.70} & \textbf{79.56} & \textbf{46.35} & \textbf{40.62} & \textbf{88.52} & \textbf{32.44} & \textbf{39.53} & \textbf{84.63} & \textbf{33.97} \\\hline
        \bottomrule[1pt]   
      \end{tabular}   
      \begin{tablenotes}
        \item[*] The \textbf{bold entries} represent the best performance in each row.
      \end{tablenotes}}   
    \end{threeparttable} 
  \end{table*}

\section{Experiments}
To evaluate the effectiveness of the HWDNet model, we carried out extensive evaluations with the state-of-the-art (SOTA) works on our proposed UCM-VeID dataset.
In this section, we firstly introduce the comparison results of the proposed method and SOTA works.
Furthermore, comprehensive ablation experiments have also been conducted to investigate the effectiveness of different parts in our method.

\subsection{Experimental Settings}
Following, it mainly introduces the evaluation metrics and implementation details.

\textbf{Evaluation metrics.}
Following \cite{AncongWu2017}, cumulative matching characteristic (CMC) curve and the mean average precision (mAP) are adopted for evaluation.
CMC is mainly used to evaluate the rank-k accuracy of the matching results, where rank-1, rank-5, rank-10 scores are employed in our experiments.
mAP reflects the average performances about retrieval results of all query samples.

\textbf{Implementation Details.}
We implement our model on deep learning framework PyTorch with one Nvidia GeForce RTX 3090 GPU.
As mentioned in Section 4, a ResNet50 network pre-trained on ImageNet dataset is applied as our backbone for the two-stream structure.
The first two residual blocks are linearly related between two modalities, while the last three ones share the same parameters. 
A batch normalization layer is added after shared network layers for latent feature normalization and modality gap elimination.
At the training stage, we randomly sample 12 identities with 48 RGB images and 48 IR images as the inputs at each step, which are resized as $256\times180$.
The adopted data argumentation manners are similar to \cite{Bi-Directional2020} and the SGD optimizer is utilized with 0.9 momentum parameter.
The training epoch is set to 100 and the initial learning rate is 0.01.
The margin parameter in triple loss is fixed as 0.5.
Finally, we utilize orientation classifier loss to restrict orientation-relevant features and the combination of ID loss, triple loss and similarity enforcement loss to guide orientation-invariant feature learning.
Notably, only orientation-invariant features are used for Re-ID task.

\subsection{Comparison with State-of-the-art Methods}
To confirm the validation of our proposed method, we implement the SOTA works based on two-stream network with open source codes in this subsection for comparison, including the baseline, AGW \cite{AGW2020}, DDAG \cite{DDAG2020}, CGRNet \cite{Y.Feng2021}, MMD-ReID \cite{Jambigi2021}, DGTL \cite{H.Liu2021} and LbA \cite{H.Park2021}.
Furthermore, we adopt two testing modes: single-shot mode and multi-shot mode.
Single-shot mode means that only one sample in each ID is picked out to establish the gallery set.
While in multi-shot mode, multiple samples in each ID are selected.
At the same time, we set up two searching modes.
When the query contains RGB images and the gallery consists of IR images, the searching mode is denoted as RGB2IR, conversely, the searching mode is expressed as IR2RGB.
As listed in the Table \ref{tab:C1}, HWDNet method illustrates the superiority in comparison with the SOTA methods on UCM-VeID dataset.
Specifically, for single-shot mode, HWDNet achieves $31.57\%$ rank-1 accuracy and $46.09\%$ mAP in IR2RGB searching mode respectively, with an improvement of $5.33\%$ and $3.9\%$ over the baseline.
As for multi-shot mode, HWDNet achieves $40.62\%$ rank-1 accuracy and $32.44\%$ mAP in IR2RGB searching mode.
The corresponding improvement over the baseline are $8.03\%$ and $2.26\%$.
Similarly, our method also has the best performance in RGB2IR searching mode.
These results also indicate that the SOTA are insufficient to cross-modality vehicle Re-ID task. 
The reason accounted for this phenomenon is that lots of tricks designed for person targets may not be useful in vehicle Re-ID task.

\subsection{Ablation Study}
Aiming at evaluate the validation of each component of the proposed method, some ablation studies are conducted on the proposed dataset. 
More concretely, to evaluate the influence of some specific modules, the remainder ones would be removed from the proposed method. 
To make a fair comparison, the overall experimental settings remain invariable.
The corresponding results with IR2RGB searching mode are listed in Table \ref{tab:C2}.

\begin{table*}[!t]
  \centering
  % \scriptsize
  \small
  \renewcommand{\arraystretch}{1.2}
  \begin{threeparttable}
  \caption{The influence of each component on the performance of the proposed HWDNet in single-shot mode on UCM-VeID datasets. The rank-k (\%) and mAP (\%) are demonstrated.}
  \label{tab:C3}
  \tabcolsep=4pt
  %\begin{minipage}{20cm}
  \setlength{\tabcolsep}{3mm}{
  \begin{tabular}{cccc|cccc|cccc}
  \toprule[1pt] %\toprule define the top line width
  \multirow{2}{*}{Baseline} &\multirow{2}{*}{$\mathcal{L}_{wr}$} &\multirow{2}{*}{$\mathcal{L}_R$} &\multirow{2}{*}{$\mathcal{L}^{'}_C$} &\multicolumn{4}{c}{Single-shot} \vline &\multicolumn{4}{c}{Multi-shot} \\
  \cline{5-12}
  &  &  &  &rank1 &rank10 &rank20 &mAP  &rank1 &rank10 &rank20 &mAP \\  \hline
  \color{red}\checkmark &  &  &  &26.24 &75.01 &86.22 &42.19 &32.59 &82.21 &92.95 &30.18\\
  \color{red}\checkmark & \color{red}\checkmark &  &  &27.59 &74.62 &87.16 &42.68 &33.71 &81.36 &91.36 &30.25\\
  \color{red}\checkmark &  & \color{red}\checkmark &  &27.96 &77.55 &88.74 &43.99 &34.91 &84.22 &92,41 &32.23\\
  \color{red}\checkmark &  &  & \color{red}\checkmark &28.11 &77.91 &89.74 &43.98 &34.31 &84.19 &93.43 &31.12\\
  \color{red}\checkmark & &\color{red}\checkmark &\color{red}\checkmark  &29.55 &\textbf{79.59} &\textbf{90.97} &45.62 &36.32 &83.58 &93.45 &32.44\\
  \color{red}\checkmark &\color{red}\checkmark &\color{red}\checkmark &\color{red}\checkmark &\textbf{31.57} &78.85 &90.19 &\textbf{46.09} &\textbf{40.62} &\textbf{88.52} &\textbf{96.47} &\textbf{33.53}\\
  \bottomrule[1pt]
  \end{tabular}
  }
  \end{threeparttable}
\end{table*}

\textbf{Effectiveness of weight restrainer.}
We try to eliminate cross-modality discrepancy by designing a weight restrainer to enhance the intra-modality feature representations.
Compared with the baseline, our weight restrainer effectively improves the model performance for rank-1/mAP on both testing modes respectively. 
These experimental results suggest that the restriction of specific modality weights helps to learn the shared features between modalities in shallow layers, and can effectively reduce cross-modality discrepancy.

\textbf{Effectiveness of orientation-invariant feature separation.}
The OIFS task is devised to separate out the features containing orientation information. 
The results are demonstrated in the line 3 of Table \ref{tab:C3}.
According to these results, we can find that, with the help of OIFS, the proposed method outperforms the baseline by 1.72\%/1.8\% for rank-1/mAP in single-shot mode. 
As for the multi-shot mode, the performance gain of our proposed method are 2.32\%/2.05\% in rank-1/mAP.
This phenomenon validates the effectiveness of our devised OIFS stage. 

\textbf{Effectiveness of orientation-invariant feature refining.}
Since the OIFR strategy is developed to further eliminate the semantic gap raised by the cross-modality discrepancy, it has great impact on the final performances of our proposed method.
Therefore, it is necessary to confirm its effectiveness and the corresponding experimental results are listed in the line 4 of Table \ref{tab:C3}. 
These results demonstrate that the performance promotions derives from OIFR strategy under single-shot and multi-shot modes are 1.87\%/1.79\% and 1.72\%/0.94\% in rank-1/mAP, respectively. 
This proves that OIFR strategy indeed contributes to the capacity of our proposed method for vehicle Re-ID task.

\textbf{Effectiveness of decoupling feature learning.}
The decoupling feature learning task is composed of two subtasks: OIFS task and OIFR task.
In order to evaluate the effectiveness of decoupling feature learning task, we performed both OIFS and OIFR tasks at the same time.
As shown in the $5\text{-th}$ line of Table \ref{tab:C2}, the performance of decoupling feature learning task is better than that of any subtask, suggesting that two subtasks can benefit each other in the training stage to obtain a more discriminative orientation-invariant feature.

\par 
Finally, the performances of different combinations of each component are evaluated and listed in the line 5 to line 6 in Table \ref{tab:C3}.
On the basis of these results, it is effortless to infer that, with the incorporation of different proposed modules into the baseline framework, an impressive performance promotion is available on the proposed UCM-VeID dataset. 
In conclusion, not only each independent module but also the whole proposed framework can be beneficial to the cross-modality vehicle Re-ID task.

\begin{figure}[h]                  
  \centering                    
  \includegraphics[width=\linewidth]{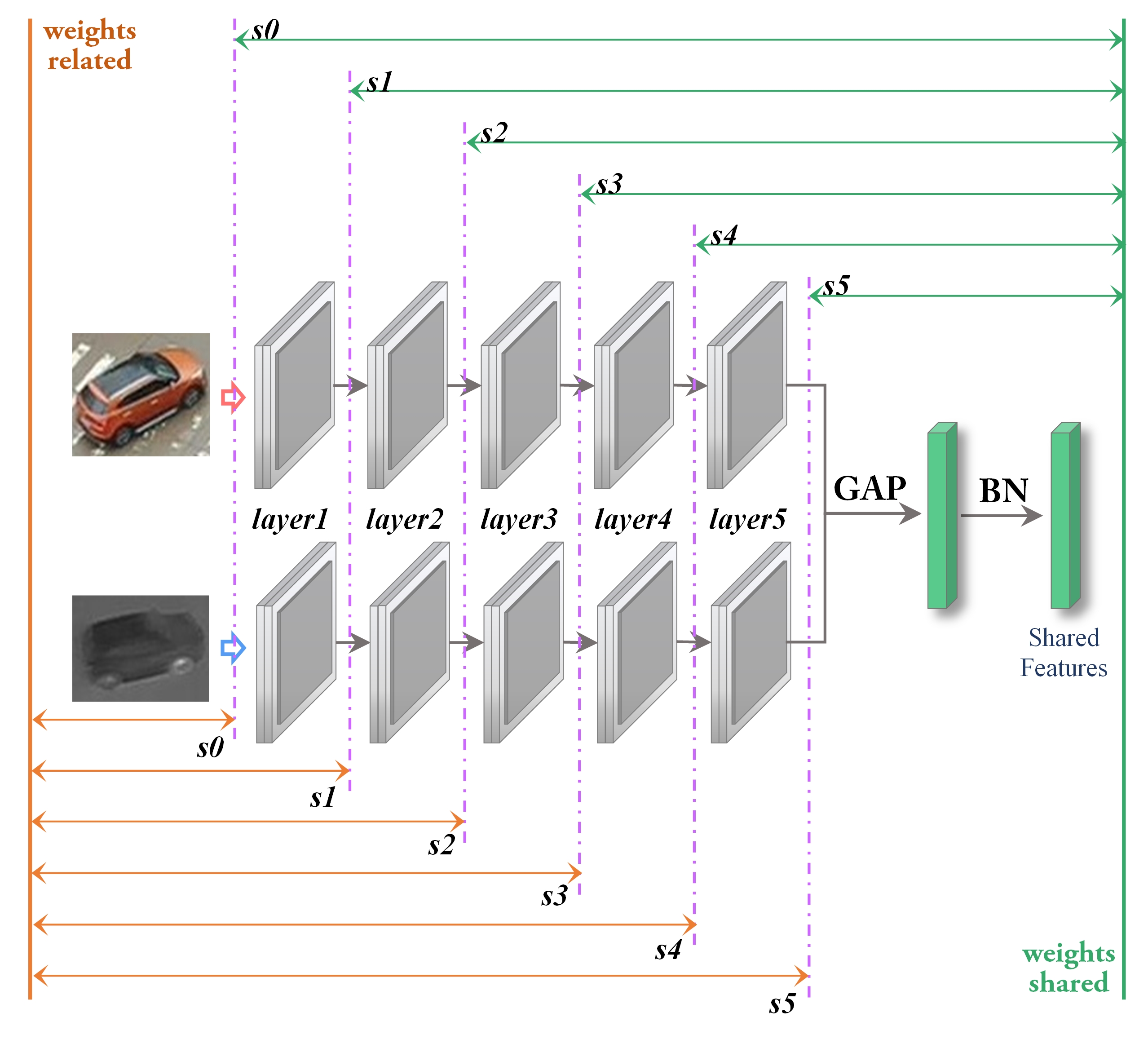}                  
  \caption{Illustration of the baseline network according to different parameters split strategy based on ResNet50 model, where $stage=si,i=\{0,1,2,3,4,5\}$ denotes the layers whose weights should be related.}                  
  \label{fig:C11}    
\end{figure}

\begin{table}[h]
  % \small
  \centering   
  % \scriptsize  
  \footnotesize 
\begin{threeparttable} 
  \renewcommand{\arraystretch}{1.5}   
  \caption{Analysis of the graph-level relation modeling on UCM-VeID dataset. The rank-k (\%) and mAP (\%) are demonstrated.} 
  \label{tab:C4}
  \setlength{\tabcolsep}{1.3mm}
  \begin{tabular}{ |c|c|c|c|c|c|c|c| }
    % \toprule[1pt]
    \hline 
    Stage    & Related layer     & Shared  layer     & Rank1      & Rank10     & Rank20     & mAP      \\ \hline
    s0       & none               & all                   & 30.19       & 82.69      & 93.91      & 29.10      \\ \hline
    s1       & layer\{1\}          & layer\{2-5\}         & 34.88       & 84.34      & 93.74      & 30.60      \\ \hline    
    s2       & layer\{1-2\}        & layer\{3-5\}         &\textbf{40.62} &\textbf{88.52} &\textbf{96.47} &\textbf{32.44}       \\ \hline
    s3       & layer\{1-3\}        & layer\{4-5\}         & 37.18       & 83.54      & 92.29      & 31.23      \\ \hline
    s4       & layer\{1-4\}        & layer\{5\}           & 34.35       & 84.06      & 94.08      & 30.41      \\ \hline
    s5       & all              & GAP+BN                  & 28.67       & 80.19      & 92.46      & 27.57      \\   \hline 
    % \bottomrule[1pt]
  \end{tabular}
\end{threeparttable} 
\end{table}

\subsection{Parameter Analysis}
\textbf{Evaluation of different stages for weight restrainer.}
In this paper, the key point of the two-stream network is to determining which layers of the network are related and which layers of the network are shared.
Based on the baseline, we optionally build the following network structures with the ResNet50 model as shown in the Fig. \ref{fig:C11}.
In IR2RGB searching mode and multi-shot testing mode, the results of different backbones splits on UCM-VeID dataset are listed in Table \ref{tab:C4}, from which we can observe that:
s5 means that all five layers weights of two branch networks in ResNet50 are related to each other, which obtains the worst performance on UCM-VeID dataset.
Although, weight restrainer could prevent different modalities representations from being too far from each other, it is insufficient to learn the unite latent space of different modalities by weight restrainer and shared fully connected layer.
s0 makes the two branch networks share all weights to learn the representations of different modalities in the same latent space, which leads to a unsatisfied performance.
s0 treats both RGB and IR inputs equally, which loses modality-specific information, such as color information and intensity information.
However, compared with the results of s5, it is enough to prove that parameter sharing is the powerful operation to solve the cross-modality problem.
Furthermore, s1, s2 and s3 achieve comparable performances.  
Overall, s2 can achieve the best performance, which only sets layer0 and layer1 as the weight related layers while others parameters sharing.

\begin{table}[t!]
  % \small
  \centering   
  % \scriptsize  
  \footnotesize 
\begin{threeparttable} 
  \renewcommand{\arraystretch}{1.5}   
  \caption{Decoupling structure. The rank-k (\%) and mAP (\%) are demonstrated.} 
  \label{tab:C5}
  \setlength{\tabcolsep}{3mm}
  \begin{tabular}{ |c|c|c|c|c| }
    % \toprule[1pt] 
    \hline
    structure   & Rank1      & Rank10     & Rank20     & mAP      \\ \hline
    Feature split   &\textbf{40.62} &\textbf{88.52} &\textbf{96.47} &\textbf{32.44}       \\ \hline
    Feature subtraction &36.18 &84.57 &93.99 &31.76      \\ \hline    
    Feature prediction &34.80 &81.84 &92.20 &30.59 \\ \hline   
    % \bottomrule[1pt]
  \end{tabular}
\end{threeparttable} 
\end{table}
\textbf{Evaluation of different decoupling structures.}
In this paper, the decoupling strategy has been comprehensively investigated and three decoupled structures are presented. 
In order to figure out which structure refers to the best structure, we have evaluated the performances of our proposed method under different decoupling structures. 
To conduct a comprehensive comparison, Rank1, Rank10, Rank20 and mAP are utilized as the metrics and the correspond results under IR2RGB searching mode and multi-shot testing mode are listed in Table \ref{tab:C4}.
According to the results in Table \ref{tab:C4}, we can make a conclusion that all the proposed decoupling structures have positive impact on the final performance of our proposed method. 
Furthermore, the best structure refers to feature split structure, since the Rand1, Rank10, Rank20 and mAP are 36.36\%, 86.90\%, 96.03\% and 31.89\% respectively. 
This phenomenon indicates that more network parameters in decoupling structure may have negative influence on the performances of our proposed method.
Meanwhile, it also validates that the devised OIFS and OIFR strategies are competent with the feature decoupling task.

% \begin{figure}[h]                  
%   \centering                    
%   \includegraphics[width=\linewidth]{fig8.jpg}                  
%   \caption{Decoupling Results with different output dimensions. (a) Rank-1 results (b) mAP results }                 
%   \label{fig:C8}    
% \end{figure}
% \textbf{Evaluation of different decoupled feature dimensions.} 
% In order to further verify the performance of the decoupling structure proposed in this paper, a comparison experiment was conducted with the best decoupling structure under different feature dimensions.
% As illustrated in Fig \ref{fig:C8}, the best feature dimension refers to 2048, which achieve the best accuracy of rank1 as $40.67\%$ and the highest accuracy of mAP as $35.79\%$.
% This phenomenon indicates that low-dimensional features may not be adequate to represent the target, while high-dimensional features may be too miscellaneous.
% The features of 2048 dimension is the most suitable for the characterization of multi-modal vehicle targets in remote sensing scenarios.

\begin{table}[h]
  % \small
  \centering   
  % \scriptsize  
  \footnotesize 
\begin{threeparttable} 
  \renewcommand{\arraystretch}{1.5}   
  \caption{Centroid construction. The rank-k (\%) and mAP (\%) are demonstrated.} 
  \label{tab:C6}
  \setlength{\tabcolsep}{3mm}
  \begin{tabular}{ |c|c|c|c|c| }
    % \toprule[1pt] 
    \hline
    Centroid   & Rank1    & Rank10     & Rank20     & mAP      \\ \hline
    single-modality       & 35.24       & 85.25      & 93.52      & 31.91      \\ \hline   
    cross-modality        &\textbf{40.62} &\textbf{88.52} &\textbf{96.47} &\textbf{32.44}       \\  \hline  
    % \bottomrule[1pt]
  \end{tabular}
\end{threeparttable} 
\end{table}
\textbf{Evaluation of different statistical centroids.}
We evaluated two different statistical centroids used to guide orientation-invariant feature learning, and the results are shown in Table \ref{tab:C6}. 
We observe that the average operation based on a cross-modality statistical centroid has better results than single-modality, suggesting that cross-modality statistical centroid can effectively reduce modality discrepancy. 
% In addition, the performance is further improved by cross-modality centroid in Mahalanobis space, which takes feature variations and unit size inconsistencies of different features into account.

\begin{figure}[h]                  
  \centering                    
  \includegraphics[width=\linewidth]{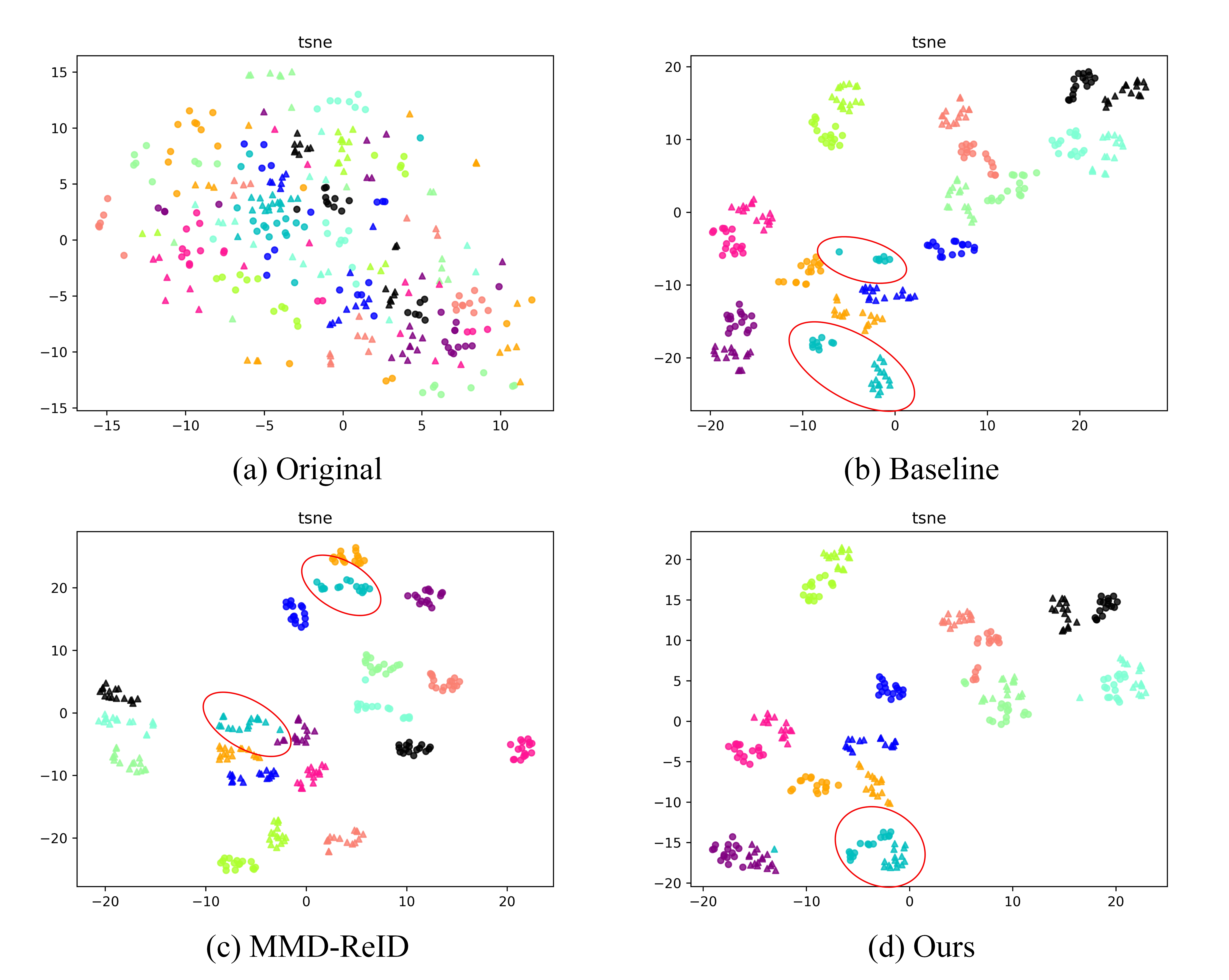}                  
  \caption{Feature distribution. (a-d) show the distribution of feature embeddings in the 2D feature space, where circles and triangles in different colors denote RGB and IR modalities with different identities.}                  
  \label{fig:C9}    
\end{figure}
\begin{figure*}[!t]                  
  \centering                    
  \includegraphics[width=\linewidth]{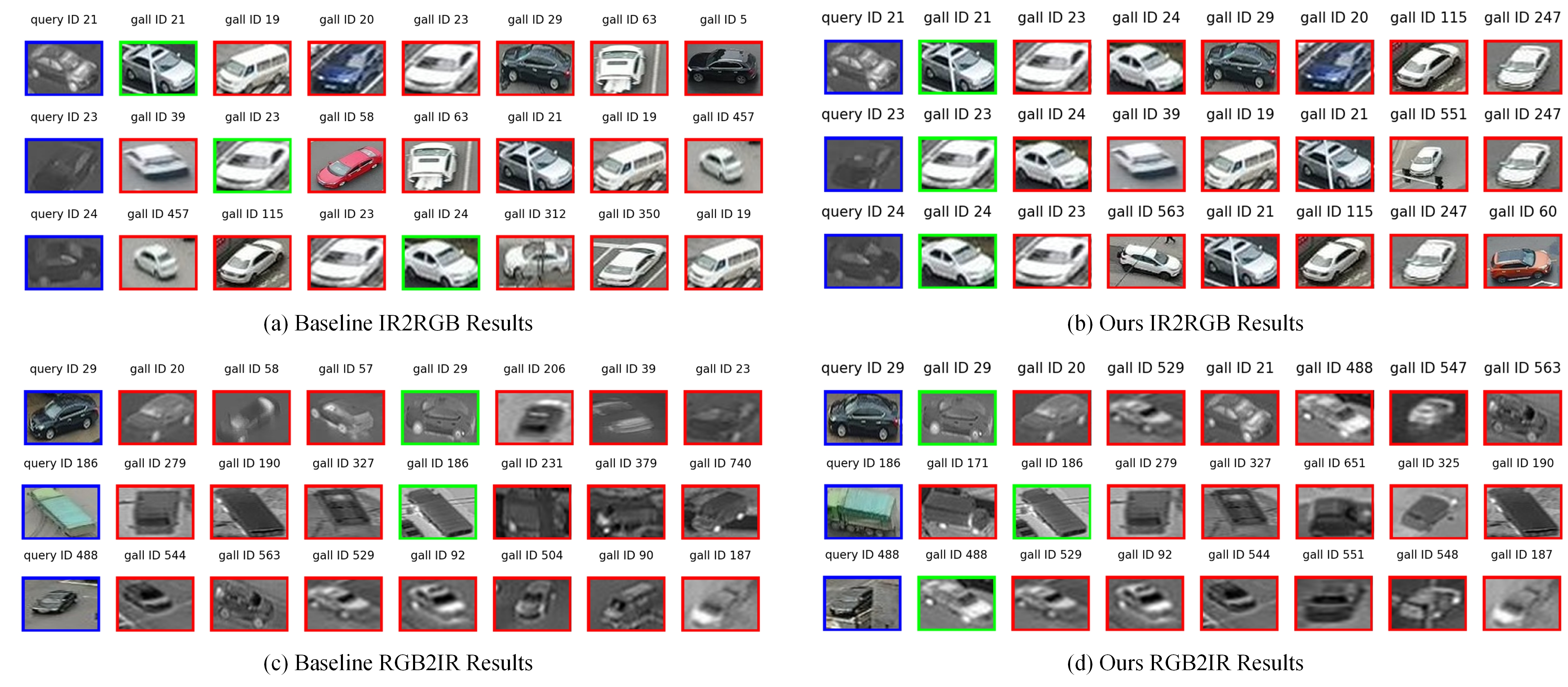}                  
  \caption{Some Rank-7 retrieval results obtained by the proposed baseline and CM-RIFL on our ID dataset UAV-CM-VeID. (a-d) are in the single-shot testing mode. The images with green borders belongs to the same identity as the given query which has blue border, red is opposite.}                  
  \label{fig:C10}    
\end{figure*}
\subsection{Visualization}
\textbf{Feature distribution.}
In Fig. \ref{fig:C9}, we use t-SNE to conduct the 2D feature spaces to visualize the latent feature distributions. 
Like the target circled in red in Fig. \ref{fig:C9}, it can be found that in our method, RGB and IR samples with same ID are closer in the feature space.
While the same ID samples are farther away in the feature space with baseline and suboptimal model both in inter-modality and intra-modality.
This figure indicates that, compare to the baseline and suboptimal model, our proposed method achieves a better performance on discriminating feature representations of the same ID, while bridging the cross-modality semantic gap. 
Moreover, our proposed method can also effectively enhance the compactness of inter- and intra-modality features of the same identity, which can significantly improve the accuracies of Re-ID results. 

\textbf{Retrieval result.}
To comprehensively illustrate the superiority of our proposed method, we further visualize the retrieval results of our proposed method and the baseline, achieved by performing multiple queries on our proposed dataset. 
The top-7 ranking results in the single-shot testing mode are selected and demonstrated in Fig.\ref{fig:C10}. 
It is easy to find that more positive samples are selected by our proposed method and the rank of these selected samples  in the most top positions, whether in IR2RGB or RGB2IR mode.
At the same time, the method in this paper can select positive samples with different orientation from the query target, as shown in the last line in Fig.\ref{fig:C10} (d).

\section{Conclusion}
In this paper, we contribute the first UAV cross-modality vehicle dataset together with a novel HWDNet for cross-modality vehicle Re-ID task.
HWDNet contains a feature extractor and a decoupling structure, devoting to tackle cross-modality discrepancy and orientation discrepancy challenges respectively. 
The feature extractor is a two-stream network with a well-designed weights constrainer, which could reserve low-level shared semantic to eliminate cross-modality discrepancy.
% prevent the weights of different modalities getting too far
Meanwhile, we investigate three decoupling structures to effectively separate the shared features into two parts.
Along with the best decoupling structure, we design two stages of learning orientation-invariant features. 
In the first stage, orientation-invariant features are separated by orientation classification tasks, and in the second stage, orientation-invariant features are refined by feature similarity enforcement operation with cross-modality statistical centroid.
HWDNet method outperforms the SOTA methods on UCM-VeID dataset and shows great robustness.
We believe that our work will benefit the vehicle Re-ID community, and we will also improve the UCM-VeID dataset in terms of target motion state and number of camera platforms in the next step.

\bibliographystyle{IEEEtran}
\bibliography{ref.bib}

\end{document}